\documentclass[a4paper,fleqn]{cas-dc}

\usepackage[numbers]{natbib}
\usepackage{amsmath}
\usepackage{subfigure}
\usepackage{fixltx2e}
\usepackage{booktabs}
\newcommand{\tabincell}[2]{\begin{tabular}{@{}#1@{}}#2\end{tabular}}
\newcommand{\tablefont}{\fontsize{8pt}{\baselineskip}\selectfont}
\usepackage{multirow}
\usepackage{url}
\usepackage{bm}
\newcommand{\jq}[1]{\textcolor{black}{#1}}

\def\tsc#1{\csdef{#1}{\textsc{\lowercase{#1}}\xspace}}
\tsc{WGM}
\tsc{QE}
\tsc{EP}
\tsc{PMS}
\tsc{BEC}
\tsc{DE}
\begin{document}
\let\WriteBookmarks\relax
\def\floatpagepagefraction{1}
\def\textpagefraction{.001}

% Short title
\shorttitle{A Robust Scheme for 3D Point Cloud Copy Detection}    

% Short author
\shortauthors{Jiaqi~Yang et~al.}  

% Main title of the paper
\title [mode = title]{A Robust Scheme for 3D Point Cloud Copy Detection}  

% Title footnote mark
% eg: \tnotemark[1]
% \tnotemark[<tnote number>] 

% Title footnote 1.
% eg: \tnotetext[1]{Title footnote text}
% \tnotetext[<tnote number>]{<tnote text>} 
% \author{Jiaqi~Yang,~%\IEEEmembership{Member,~IEEE,} 
%         Xuequan~Lu,~\IEEEmembership{Member,~IEEE,}
%         Wenzhi Chen,~\IEEEmembership{Member,~IEEE}

% \thanks{Manuscript received April, 2021; revised ZZ, 2021. (Corresponding author: Xuequan Lu, Wenzhi Chen.)}    
% \IEEEcompsocitemizethanks{
% \IEEEcompsocthanksitem J. Yang and W. Chen are with the College of Computer Science and Technology, Zhejiang University, Hangzhou 310027, China
% (e-mail: yangjiaqi@zju.edu.cn; chenwz@zju.edu.cn).
% \IEEEcompsocthanksitem X. Lu is with School of Information Technology, Deakin University, Australia.
% E-mail: xuequan.lu@deakin.edu.au.
% }% <-this % stops an unwanted space

\author[1]{Jiaqi~Yang}[style=chinese]
\ead{yangjiaqi@zju.edu.cn}

\author[2]{Xuequan~Lu}[style=chinese,orcid=0000-0003-0959-408X]
\ead{xuequan.lu@deakin.edu.au}
\ead[url]{http://www.xuequanlu.com}

\author[1]{Wenzhi~Chen}[style=chinese]
\cormark[1]
\ead{chenwz@zju.edu.cn}
% First author
%
% Options: Use if required
% eg: \author[1,3]{Author Name}[type=editor,
%       style=chinese,
%       auid=000,
%       bioid=1,
%       prefix=Sir,
%       orcid=0000-0000-0000-0000,
%       facebook=<facebook id>,
%       twitter=<twitter id>,
%       linkedin=<linkedin id>,
%       gplus=<gplus id>]

% \author[<aff no>]{<author name>}[<options>]

% Corresponding author indication

% Footnote of the first author
% \fnmark[<footnote mark no>]

% Email id of the first author
% \ead{<email address>}

% URL of the first author
% \ead[url]{<URL>}

% Credit authorship
% eg: \credit{Conceptualization of this study, Methodology, Software}
% \credit{<Credit authorship details>}

\address[1]{College of Computer Science and Technology, Zhejiang University, Hangzhou 310027, China}
\address[2]{School of Information Technology, Deakin University, Australia}

% Address/affiliation
% \affiliation[<aff no>]{organization={},
%             addressline={}, 
%             city={},
% %          citysep={}, % Uncomment if no comma needed between city and postcode
%             postcode={}, 
%             state={},
%             country={}}

% \author[<aff no>]{<author name>}[<options>]

% Footnote of the second author
% \fnmark[2]

% Email id of the second author
% \ead{}

% % URL of the second author
% \ead[url]{}

% % Credit authorship
% \credit{}

% % Address/affiliation
% \affiliation[<aff no>]{organization={},
%             addressline={}, 
%             city={},
% %          citysep={}, % Uncomment if no comma needed between city and postcode
%             postcode={}, 
%             state={},
%             country={}}

% Corresponding author text
\cortext[1]{Corresponding author}

% Footnote text
% \fntext[1]{}

% For a title note without a number/mark
%\nonumnote{}

% Here goes the abstract
\begin{abstract}
Most existing 3D geometry copy detection research focused on 3D watermarking, which first embeds ``watermarks'' and then detects the added watermarks. However, this kind of methods is non-straightforward and may be less robust to attacks such as cropping and noise. In this paper, we focus on a fundamental and practical research problem:
\jq{judging whether a point cloud is plagiarized or copied to another point cloud in the presence of several manipulations (e.g., similarity transformation, smoothing).}
%judging if two given point clouds are similar (e.g., partly similar, or the same point cloud models) regardless of the categories, in the presence of attacks.
We propose a novel method to address this critical problem. Our key idea is first to align the two point clouds and then calculate their similarity distance. We design three different measures to compute the similarity. We also introduce two strategies to speed up our method. Comprehensive experiments and comparisons demonstrate the effectiveness and robustness of our method in estimating the similarity of two given 3D point clouds.
\end{abstract}

% Use if graphical abstract is present
%\begin{graphicalabstract}
%\includegraphics{}
%\end{graphicalabstract}

% Research highlights
% \begin{highlights}
% \item 
% \item 
% \item 
% \end{highlights}

% Keywords
% Each keyword is seperated by \sep
\begin{keywords}
 3D point cloud copy detection \sep 3D shapes \sep 3D watermarking \sep GMM \sep similarity \sep
\end{keywords}

\maketitle

% Main text
%-------------------------
\section{Introduction}
The copy detection of 3D geometric data has received noticeable attentions in recent years. It is of great significance to copyright 3D geometric data, especially valuable high-quality 3D models of cultural heritage. The 3D models are often obtained from a series of steps such as scanning, smoothing, hole filling, and surface reconstruction, etc. Those steps are non-trivial and could be time-consuming and cost-expensive. As a result, it is more challenging to capture 3D models than 2D images. Thus, it is very significant to value the owners' efforts and detect 3D models' copies.

In this work, we focus on a fundamental research problem which is to judge if two point clouds (a 3D representation with a set of unordered points) are similar or not (e.g., partly similar, or the same point cloud models). The problem comes from the situation in which a user uses a 3D model, and we need to judge if this model is a manipulated duplication of a ``ground-truth'' model at hand (e.g., rotated copy, noisy copy, and cropped copy). If they are judged to be highly similar, the user's model is most likely a copy, and the user may violate the copyright of that model. 

Existing methods mainly focused on borrowing the ``watermarking'' concept from digital watermarking \cite{chou2007technologies,medimegh2015survey,shukla2012watermarking}. The pipeline is very similar to digital watermarking on images: first adding watermarks and then detecting watermarks for recognition. However, the watermarking techniques depend greatly on watermarks which may be fragile to attacks like noise and cropping. For instance, if the 3D model is corrupted with considerable noise, the watermarks can hardly be extracted successfully. It is non-straightforward to add watermarks and then detect watermarks. An alternative is using 3D shape retrieval methods to compare the similarity of two 3D shapes. Nevertheless, 3D shape retrieval methods aim to search models of the same category as the query model, while our target in this work (i.e., judging if two models are similar regardless of the categories, even in the presence of attacks).

We are motivated by the above analysis and propose a novel approach for 3D point cloud copy detection. Our core idea is first to align the two point clouds and then calculate three different similarity distances, revealing the similarity degree of the two point clouds. In particular, we employ an effective point set registration algorithm - CPD \cite{myronenko2010point} for the alignment. 
We observe that the probability matrix depicts the relationship between the two point clouds and can be used as input for calculating the similarity between them. To achieve this, we perform Robust PCA on the matrix and obtain its low-rank component, representing the vital information. We design the low-rank measure like the mean of the low-rank matrix elements, which is sufficient to account for the similarity degree. In addition, we also design two other measures for speed and comparison purposes: the Kurtosis measure and the Correlation measure. Finally, we design two acceleration strategies to speed up the computation. 

The main contributions of this paper are as follows.
\begin{itemize}
    \item We present a novel technique of 3D point cloud copy detection that avoids using the watermarking concept.
    \item We design three different but effective distance measures to calculate the similarity degree between two point clouds.
    \item We conduct extensive experiments to validate our method. We compare our method with watermarking methods. Furthermore, we also compare our method with recent 3D shape retrieval approaches in the shape retrieval setting. We finally do some additional studies. Results show that our method is effective and robust in estimating the similarity of two 3D point clouds in the presence of various attacks.
\end{itemize}
\section{Related Work}
\label{sec:relatedwork}
This section mainly concentrates on the techniques that are most related to 3D point cloud copy detection, including 3D shape watermarking and 3D shape retrieval. Finally, we look back upon some related applications of the Gaussian mixture model (GMM).

\subsection{3D Shape Watermarking} 
\label{sec:3Dshapewatermarking}
Like 2D image watermarking, 3D shape watermarking adds negligible watermarks on 3D geometric models and then extracts watermarks through specifically designed algorithms. It generally includes robust and fragile schemes \cite{chou2007technologies}. The robust watermarking aims to endure malicious attacks and thus protect the copyright, while the fragile watermarking intends to check the authenticity and the integrity of the 3D models \cite{medimegh2015survey}. And we only consider the robust watermarking, which, in general, requires a sophisticated process and obscure mapping relationships to ensure the robustness and transparency of the watermark. The transparency tends to estimate the privacy of the embedded watermarks, and the robustness focuses on immutability, namely bit error rate (BER) and correlation. %\xq{[explain fragile schemes]} 

3D shape watermarking deals with the spatial domain  \cite{cho2006oblivious,amar2016euclidean,tsai2018vertex,liu2019novel,hou2017blind,liu2018blind} and the spectral domain \cite{ohbuchi2001watermarking,hamidi2017robust,hamidi2019robust,ferreira2020robust}, depending on either the geometry and connectivity information or the spectral information \cite{medimegh2015survey}. Wang et al. \cite{wang2007three} concluded that intrinsic properties of 3D meshes (i.e., chaotic topology and unpredictable sampling of 3D meshes) and diversity of malicious distortion on watermarked meshes cause 3D watermarking technology more awkward compared to the digital image processing field.

Amar et al. \cite{amar2016euclidean} quantified and deformed Euclidean distances from all vertices to the mass center of the 3D model. There is a mapping relationship between watermarks and the parity of quantization value of Euclidean distances: odd number corresponds to watermark value 0, and even number corresponds to watermark value 1. And the vertex position is modified to reduce the quantization value in the watermark embedding phase, while the watermark is determined according to the above-mentioned relationship in the watermark extraction phase.

Hamidi et al. \cite{hamidi2017robust} established a codebook with the private key, watermark, and wavelet coefficient vector (WCV) of coarsest meshes obtained through multi-resolution wavelet decomposition and reconstructed meshes using the WCV modified by the codebook. As for watermark extraction, the watermark should make the codeword in the codebook the closest to the WCV norm. After that, Hamidi et al. \cite{hamidi2019robust} improved the performance by using the vertices in saliency rather than all vertices in the mesh model.

Several additional processing techniques are devoted to watermarking for 3D printed models, which naturally introduce distortion during printing and scanning. Hou et al. \cite{hou2017blind} estimated the print axis by analyzing the layering artifact and added a sinusoidal frequency signal to the vertex coordinates calculated from the watermark. Delmotte et al. \cite{delmotte2020blind} computed the norm histogram continuously over the entire surface instead of a discrete set of vertices and shifted the mean of each bin of the norm histogram to indicate the watermark value (0 or 1). These methods eliminate the adverse effects of sampling in the scanning process.

\subsection{3D Shape Retrieval} 
\label{sec:3Dshaperetrieval}
3D shape retrieval targets to query 3D shapes which are closest to the given 3D model. In essence, 3D shape retrieval is to extract and compare the feature of a shape with that of the query model.  Different data representations and application scenarios exacerbate the complexity of shape retrieval methods \cite{xiao2020survey,li2015comparison}.  We simply cover the structure-based methods and view-based methods here.

\textbf{Structure-based approaches.} Rich surface and hidden geometric/graph structure amply depict the discrepancy among shapes. The shape descriptors mentioned in \cite{zhang2007survey} are practical in retrieval for polygon meshes and point clouds, e.g., global information \cite{osada2001matching},  local features  \cite{frome2004recognizing,tombari2010unique,salti2014shot,furuya2016deep}, Zernike moment \cite{novotni2004shape}, distribution \cite{osada2002shape,moyou2014lbo}, skeleton \cite{rezaei2018k}, topology \cite{barra20133d,som2018perturbation}. Recently, deep learning methods for 3D shape retrieval based on shape structure have been proposed. For example, Furuya et al.  \cite{furuya2016deep} introduced the DLAN to extract rotation-invariant local 3D features and aggregated these local features into global descriptors. Feng et al. \cite{feng2019meshnet}  calculated the spatial and structural descriptors of all polygon faces, and obtained the global descriptors through the combination of descriptors and neighbor aggregation operations.

\textbf{View-based approaches.} Inspired by the intuitive perception of 3D shapes, researchers proposed to convert 3D shapes to two-dimensional planes (i.e., depth maps \cite{feng20163d}, projection \cite{su2015multi,bai2017gift,huang2019deepccfv}), thus facilitating the application of mature two-dimensional retrieval techniques for 3D shapes. Su et al. \cite{su2015multi} extracted 2-dimensional features from different projection rendering views, which were computed by rotating the virtual camera for each shape. Later they aggregated these features into a global descriptor for the entire 3D shape. VGG neural networks pre-trained on ImageNet were used in their work. Instead of aggregation, Song et al. \cite{bai2017gift} matched the features of each projection with counterparts in the retrieval database one by one, applying the re-ranking component to process the matching results. Recently, there were also 3D shape retrieval studies related to metric learning. He et al. \cite{he2018triplet} proposed a triplet-center loss for view-based techniques to improve retrieval performance.

\subsection{Gaussian Mixture Model}
Gaussian mixture model (GMM) \cite{reynolds2009gaussian} is a weighted sum of $M$ Gaussian components, which interprets complex abstract problems as data fitting problems. In general, the  Expectation-maximization (EM) algorithm \cite{moon1996expectation} is applied to estimate GMM parameters. Because of its powerful capability, GMM serves in several fields extensively, such as rigid and non-rigid point set registration \cite{myronenko2007non,myronenko2010point,fan2016convex}, compressive sensing \cite{yang2014video}, speech recognition \cite{povey2010subspace}, model denoising \cite{lu2017gpf}, model reconstruction \cite{preiner2014continuous} and skeleton learning \cite{lu2018unsupervised,lu20193d}. For instance, Preiner et al. \cite{preiner2014continuous} proposed a hierarchical EM algorithm to quickly reduce the number of model points, preserving the utmost details. Lu et al. \cite{lu2017gpf} fitted GMM centroids (representing the filtered points of the noisy model) to the data (the noisy model), achieving robust feature-preserving point set filtering.

\begin{figure*}
\centering
\includegraphics[width=0.9\linewidth]{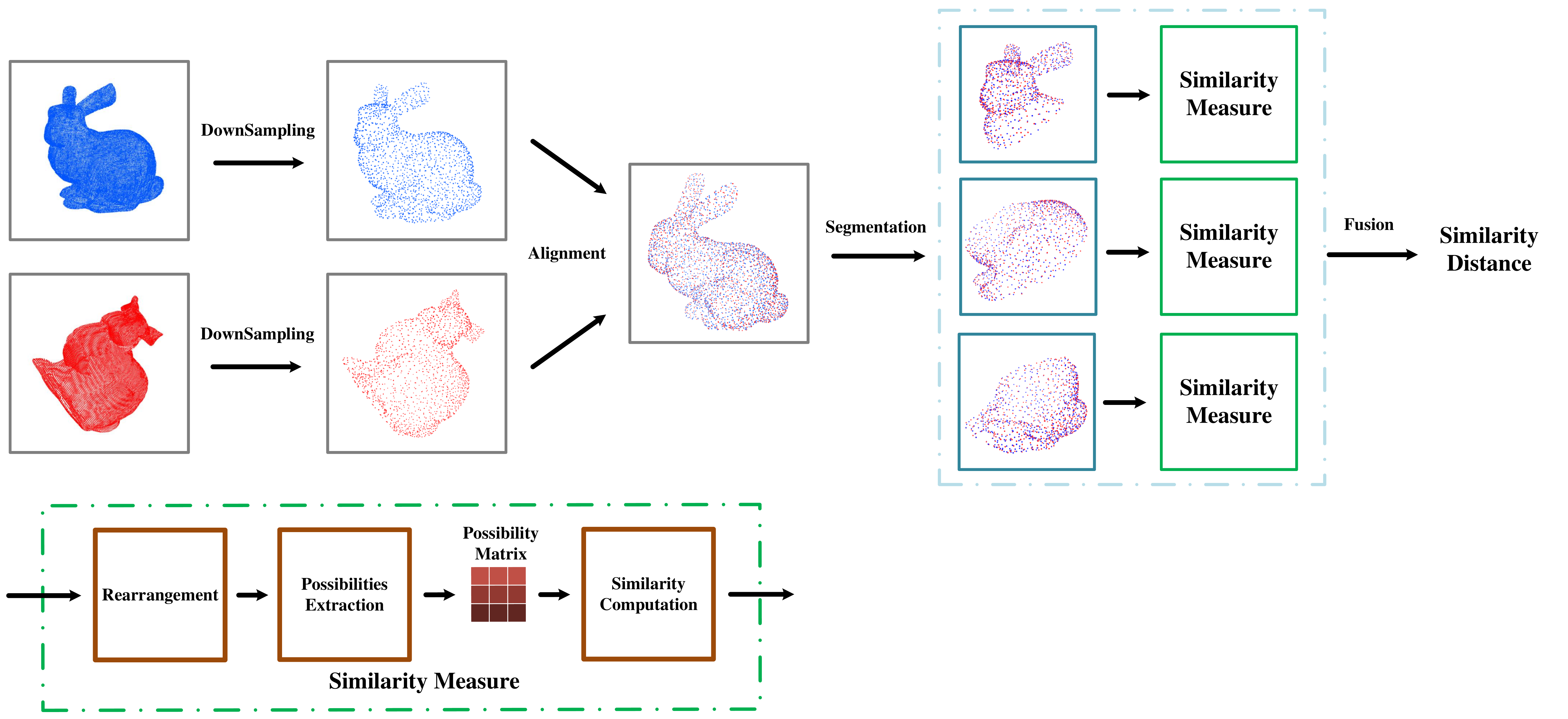}
\caption{Overview of 3D point cloud data copy detection. Two associated Bunny models are colored in red and blue.} 
\label{fig:overallflowchart}
\end{figure*}

\section{Method}
\label{sec:method}
This section explains how to detect the copies of the original 3D point cloud via our introduced method. We first give an overview of our approach and then explain each step of our method: point set registration and similarity distance measure. Finally, we introduce acceleration strategies to speed up our method.

\subsection{Overview}
Our method consists of two steps to realize the copy detection of 3D point cloud data. The first step is to align two point clouds, which ensures the fair computation of the ``distance'' between the two input models in the second step. We design three quantitative metrics to evaluate the ``distance'' between them. Also, we design strategies to speed up the computation. Figure \ref{fig:overallflowchart} illustrates the overview of the proposed method.

\subsection{Point Set Registration}
\label{sec:CPD}
Our first step is point set registration which aligns two point clouds to a similar pose. We employ the powerful CPD \cite{myronenko2010point} to achieve it. In this work, we only consider the rigid registration of two point clouds.

For two point sets $\mathbf{X}=\{\mathbf{x}_1,…,\mathbf{x}_N\}^T$ and $\mathbf{Y}=\{\mathbf{y}_1,…,\mathbf{y}_M \}^T$, we assume $\mathbf{X}$ is the sample data set generated by a GMM with each point in $\mathbf{Y}$ acting as a centroid of a Gaussian distribution. The probability of $\bm{x}_n$ is defined as Eq. \eqref{eq:px}:
\begin{equation}
\label{eq:px}
\begin{aligned}
    & p(\bm{x}_n) = \omega \frac{1}{N} + (1-\omega) \sum_{m=1}^M p(\bm{y}_m) g(\bm{x}_n|\bm{y}_m,\sigma)\\
    & g(\bm{x}_n|\bm{y}_m,\sigma) = \frac{1}
    {(2\pi\sigma^2)^{D/2}} e^{-\frac{||\bm{x}_n-\bm{y}_m||^2}{2\sigma^2}},
\end{aligned}
\end{equation}
where $p(\bm{y}_m)=\frac{1}{M}$, $D=3$ and $\sigma$ is the equal isotropic covariance of all Gaussian components. And $\omega$ is the weight of the uniform distribution which introduces an extra uniform distribution to explain noise and outliers.

$\mathbf{y}_m$ is constrained to be rigid and observes the following form:
\begin{equation}
\label{eq:ymbetween2iter}
\mathbf{y}_{m} = s\mathbf{R}\mathbf{y}_m^{ori} +\mathbf{t},
\end{equation}
where $\mathbf{y}_m^{ori}$ denotes the point of $\bm{Y}$ without any rigid transformation. $s$ is a scaling factor, $\mathbf{R}_{3\times 3}$ is a rotation matrix and $\mathbf{t}_{3\times 1}$ is a translation vector.

To achieve a GMM that best explains the relationship between the two point clouds, we rewrite Eq. \eqref{eq:px} adding rigid transformation, and minimize the negative log-likelihood $L(\bm{\theta})$.
\begin{equation}
\label{eq:negative_log_likelihood}
\begin{aligned}
L(\bm{\theta}) &= -\sum_{n=1}^{N}log\sum_{m=1}^{M}p(\bm{x}_n|\bm{\theta}) \\
p(\bm{x}_n|\bm{\theta}) &= \omega \frac{1}{N} + (1-\omega) \sum_{m=1}^M \frac{1}{M} g(\bm{x}_n|\bm{\theta})\\
g(\bm{x}_n|\bm{\theta}) &= \frac{1}{(2\pi\sigma^2)^{D/2}} 
e^{-\frac{ ||\bm{x}_n-(s\bm{R}\bm{y}_m^{ori}+\bm{t})||^2}{2\sigma^2}},
\end{aligned}
\end{equation}
where $\theta=(s,\bm{R},\bm{t},\sigma)$.

The EM algorithm is used for optimization. In the E-step, the posterior probability can be computed as: 
\begin{equation}
\label{eq:pmn}
\begin{aligned}
p_{m,n} &= p(\mathbf{y}_m | \mathbf{x}_n) \\
&= \frac
{e^{-\frac{1}{2}||\frac{\mathbf{x}_n - (s\bm{R}\bm{y}_m^{old}+\bm{t})}{\sigma}||^2}}
{\sum^M_{j=1}{e^{-\frac{1}{2}||\frac{\mathbf{x}_n - (s\bm{R}\bm{y}_j^{old}+\bm{t})}{\sigma}||^2}} + c},
\end{aligned}
\end{equation}
where $c$ is a constant independent of rigid transformation. In the M-step, we solve for $s$, $\bm{R}$, $\bm{t}$ and $\sigma$ which are discussed in detail in \cite{myronenko2010point}.
\begin{equation}
    \begin{aligned}
    &SP=\sum_{n=1}^{N}\sum_{m=1}^{M}p_{m,n}, 
    \mu_x=\frac{1}{SP} \bm{X}^T\bm{P}^T\bm{1}, 
    \mu_y=\frac{1}{SP} \bm{Y}^T\bm{P}^T\bm{1},\\
    &\hat{\bm{X}}=\bm{X}-\bm{1}\mu_x^{T},
    \hat{\bm{Y}}=\bm{Y}-\bm{1}\mu_y^{T},\\
    &\bm{U}\bm{\Sigma}\bm{V}^T=SVD(\hat{\bm{X}}^T\bm{P}^T\hat{\bm{Y}}),
    \bm{C} =d(1,\cdots,1,det(\bm{U}\bm{V}^T)),\\
    &\bm{R}= \bm{U}\bm{C}\bm{V}^T,
    s=\frac{tr(\bm{\hat{\bm{Y}}}\bm{P}^T\hat{\bm{X}}\bm{R})}{tr(\hat{\bm{Y}^T}d(\bm{P}\bm{1})\hat{\bm{Y}})},
    \bm{t}=\mu_x-s\bm{R}\mu_y,\\
    \end{aligned}
\end{equation}
where $\bm{P}$ is the matrix formed from $p_{m,n}$, and $\bm{1}$ is a column vector of all ones. $det(\cdot)$ is used to calculate the determinant and $tr(\cdot)$ is used to calculate the trace. $d(\cdot)$ is a diagonal matrix formed from the column vector and $SVD(\cdot)$ denotes singular value decomposition.

\begin{figure}[htbp]
%\vspace{-0.0cm}
\centering
\begin{minipage}[b]{0.3\linewidth}
{\label{}\includegraphics[width=1\linewidth]{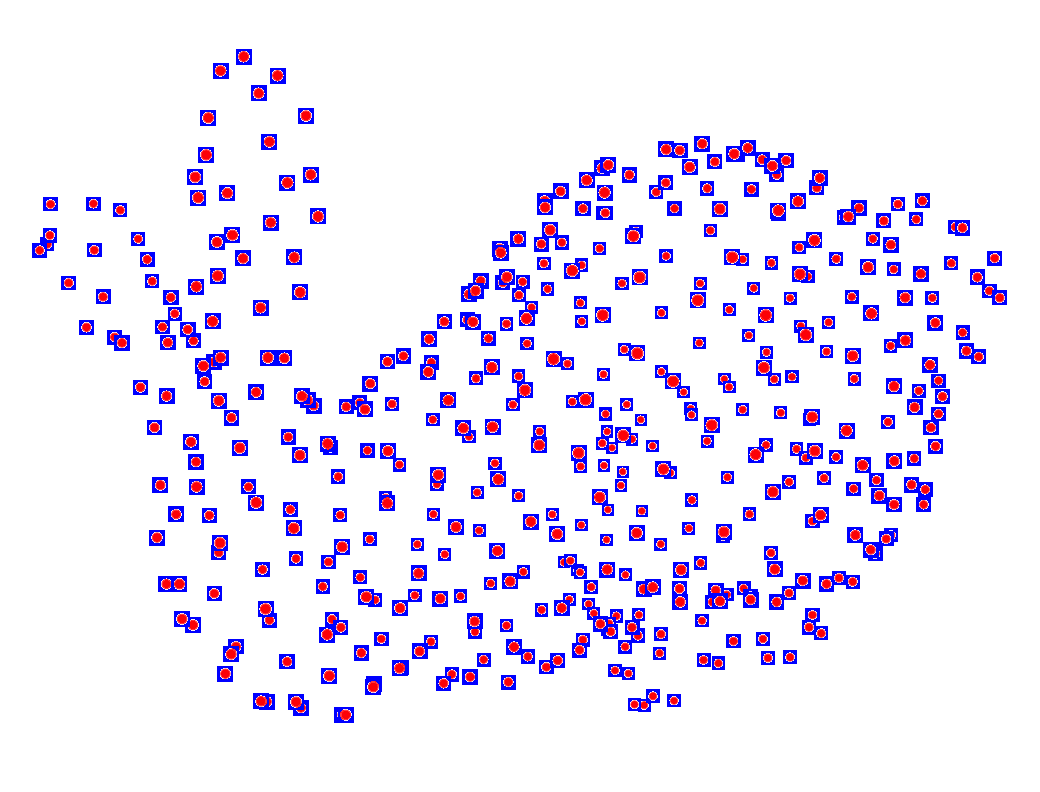}}
\end{minipage}
\begin{minipage}[b]{0.3\linewidth}
{\label{}\includegraphics[width=1\linewidth]{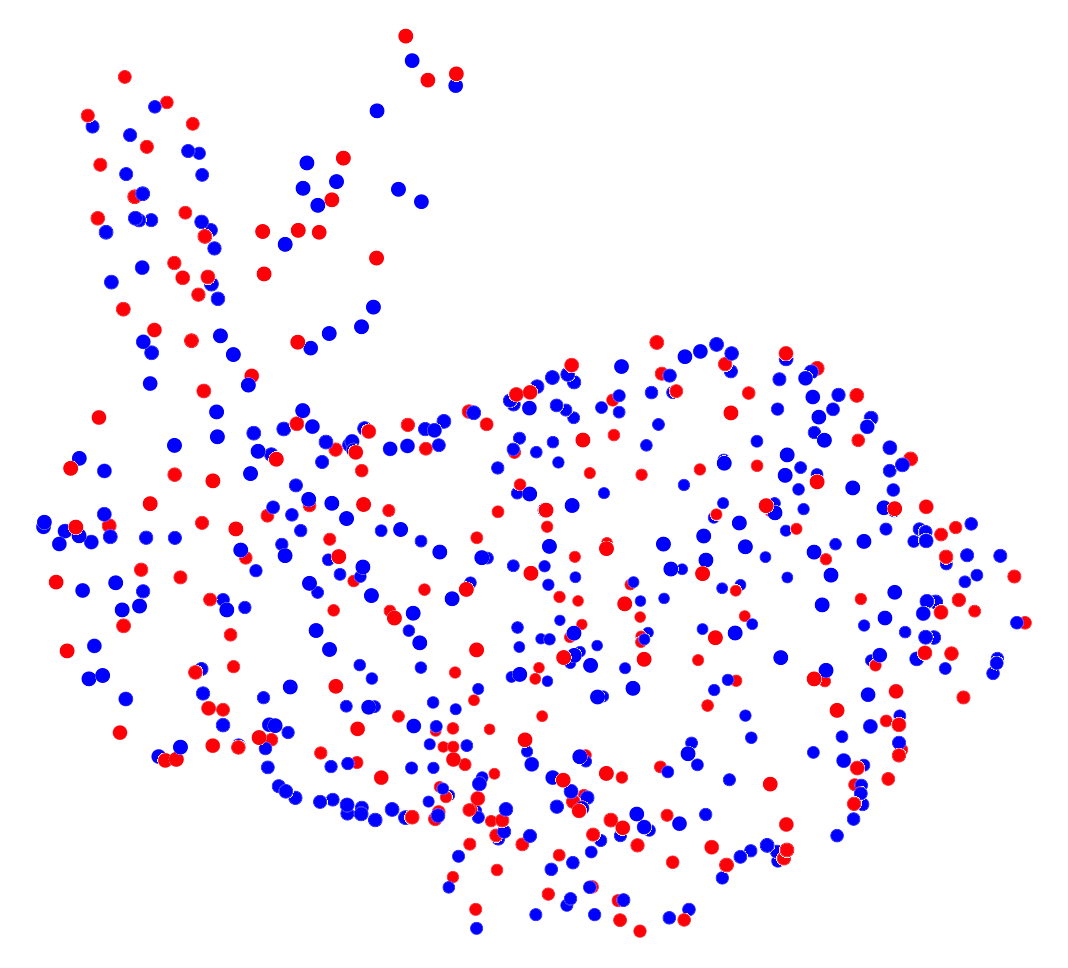}}
\end{minipage}
\begin{minipage}[b]{0.3\linewidth}
{\label{}\includegraphics[width=1\linewidth]{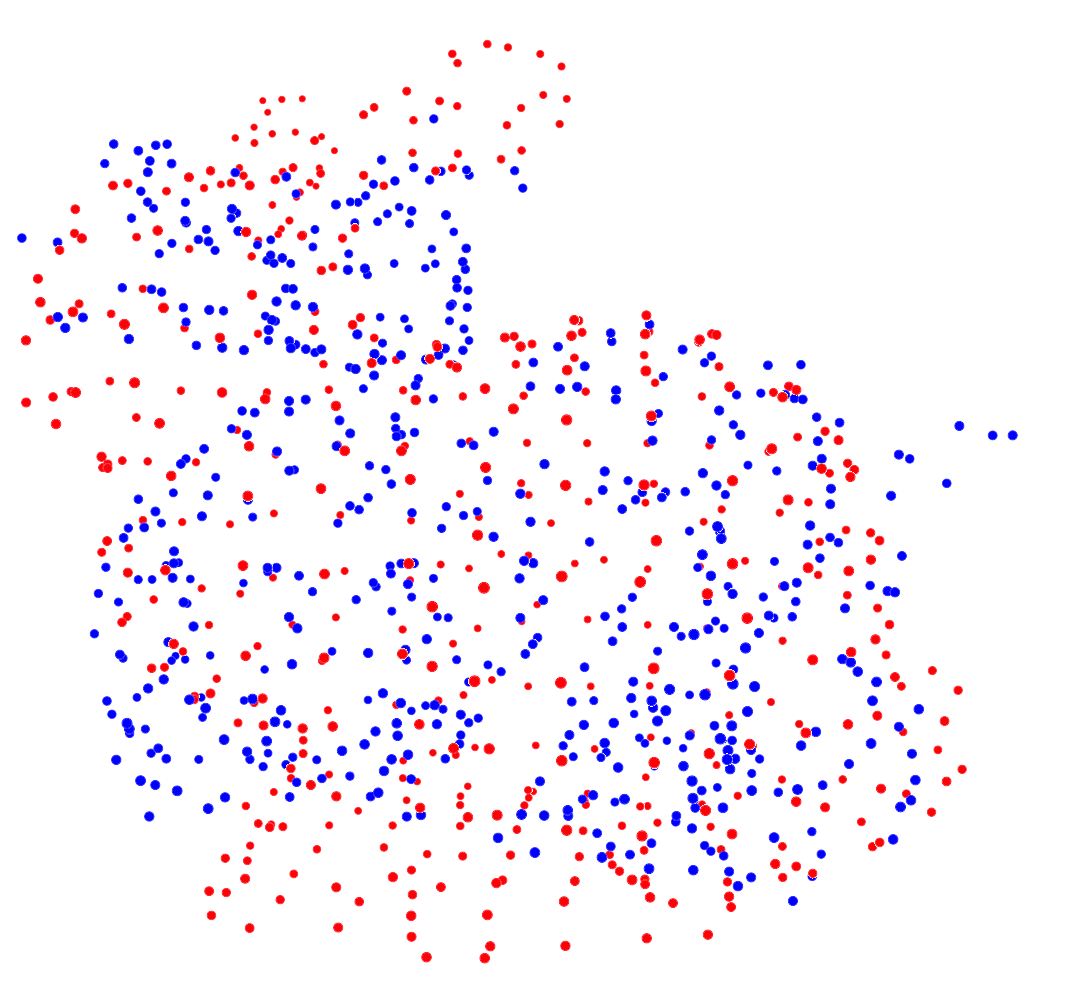}}
\end{minipage}
\begin{minipage}[b]{0.3\linewidth}
\subfigure[b453(X) b453(Y)]{\label{fig:probabilitymatrix_a}\includegraphics[width=1\linewidth]{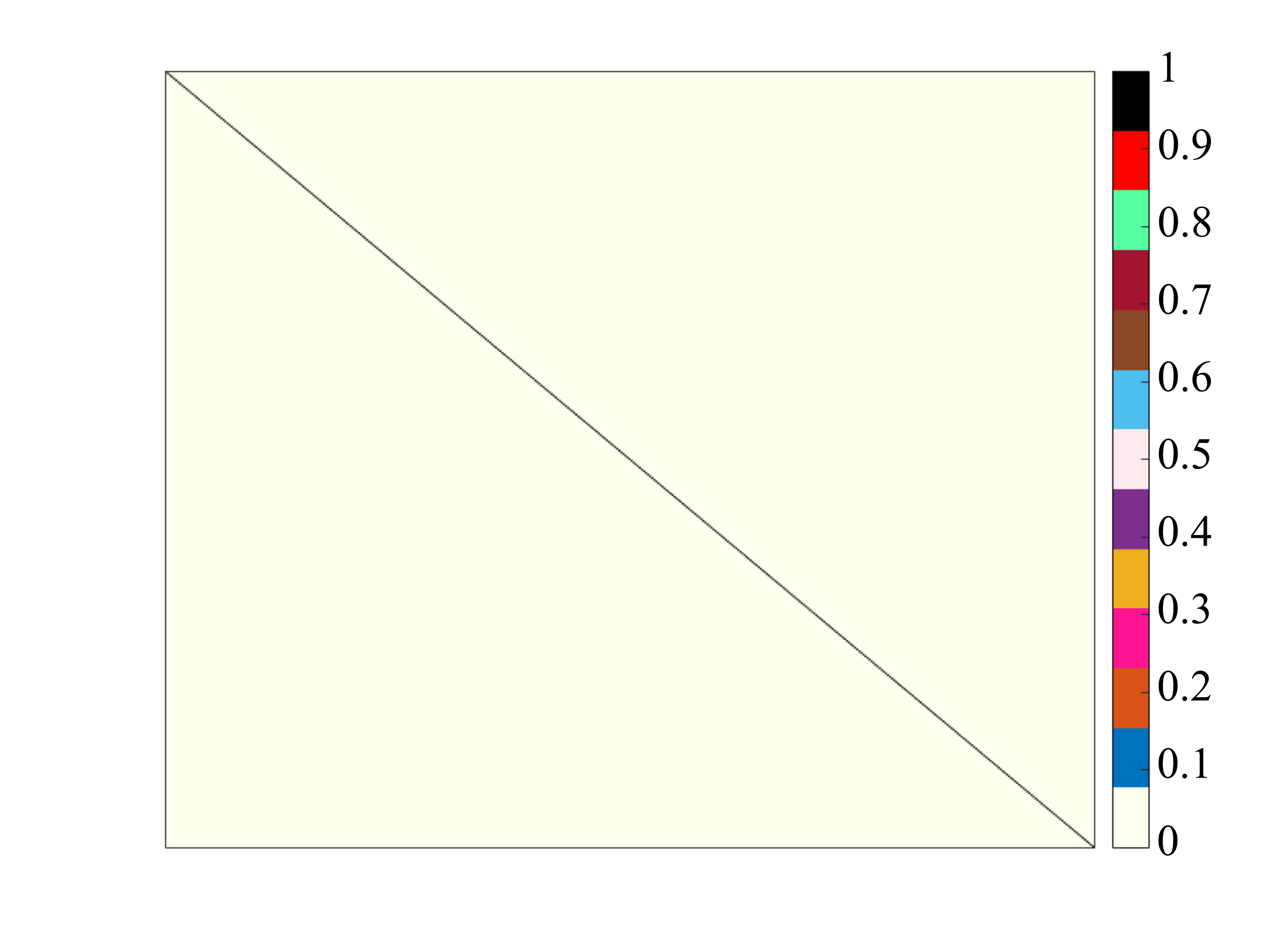}}
\end{minipage}
\begin{minipage}[b]{0.3\linewidth}
\subfigure[b227(X) b378(Y)]{\label{fig:probabilitymatrix_b}\includegraphics[width=1\linewidth]{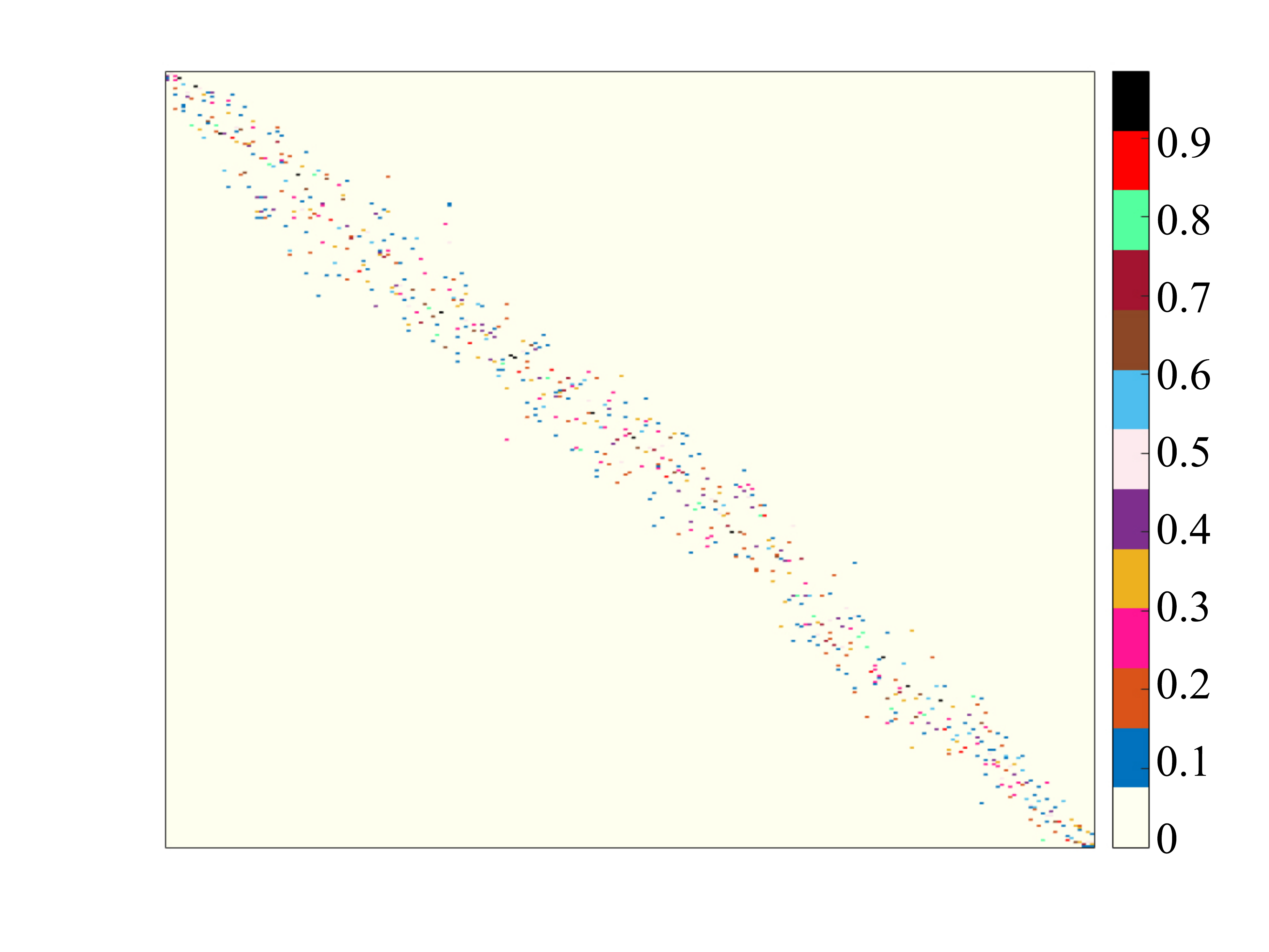}}
\end{minipage}
\begin{minipage}[b]{0.3\linewidth}
\subfigure[b453(X) d521(Y)]{\label{fig:probabilitymatrix_c}\includegraphics[width=1\linewidth]{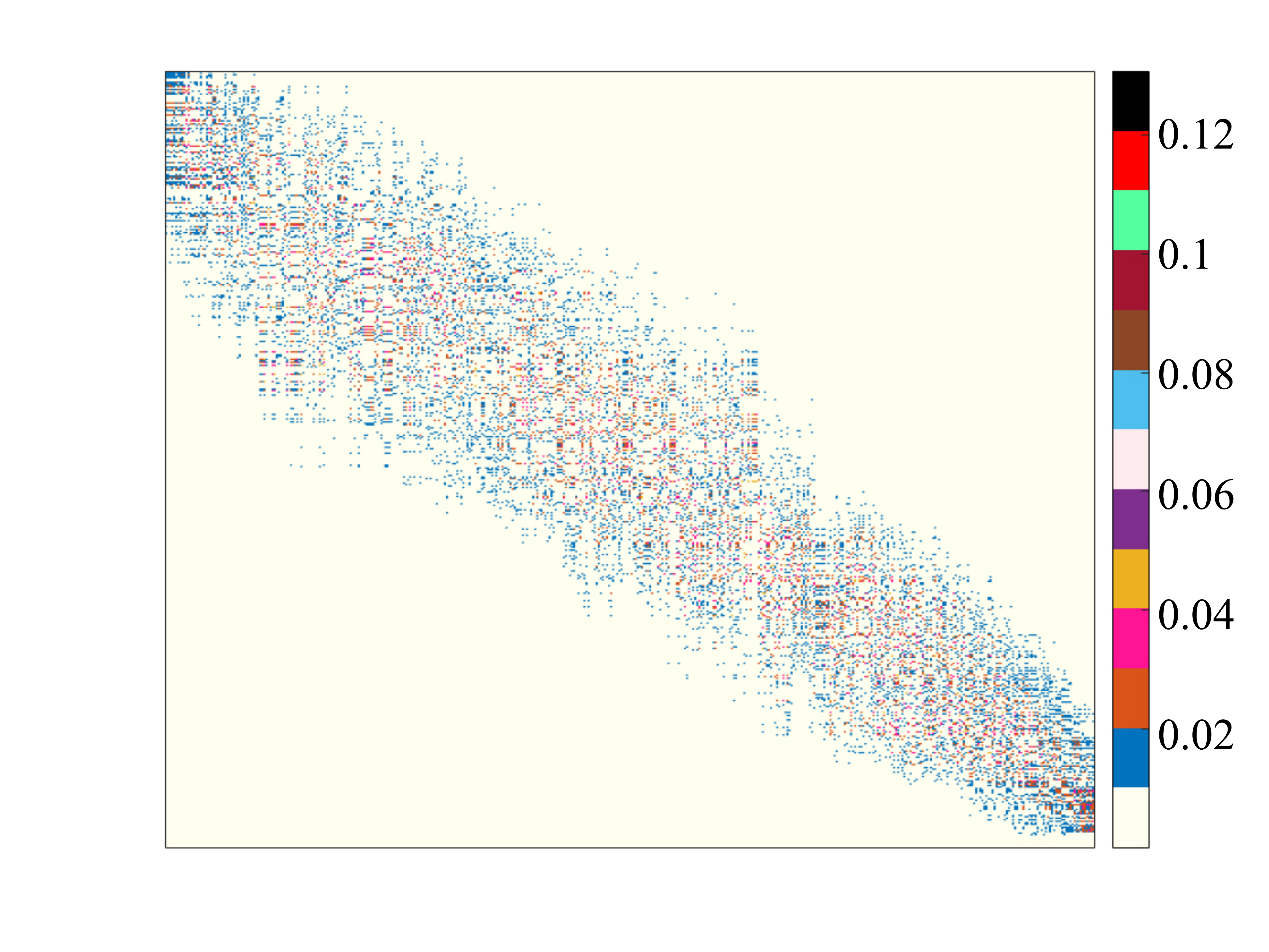}}
\end{minipage}
\caption{ The registration results (in the first row) and the probability matrices ($\bm{P}$) (in the second row) of 3 pairs of point clouds. b453 stands for the Bunny model with 453 points, d378 represents the Dragon model with 378 points, and same below. See color bars for the value ranges of each element in the probability matrix. }
\label{fig:probabilitymatrix}
\end{figure}

\subsection{Similarity Distance}
\label{sec:Identification Metrics}
This step is to estimate the distance between the two aligned point clouds. As opposed to most research that utilized point positions to compute the similarity distance, we define three distance measures based on the probability matrix ($\bm{P}$).   

Figure \ref{fig:probabilitymatrix} visualizes three probability matrices between $\bm{Y}$ and $\bm{X}$. 

Additionally, to combat the impact of disorder harmless to registration, it is necessary to reorganize points' coordinates with the rearrangement component shown in Figure \ref{fig:overallflowchart}. This component rearranges the model in the ascending order of the x, y, z coordinate values of the 3D model points (i.e., first x, then y, last z). Therefore, a colored diagonal band in Figure \ref{fig:probabilitymatrix} can express the relationship between 2 point clouds.

Apparently, the colored diagonal bands of $\bm{P}$ is tightly bounded up with the overlap of $\bm{X}$ and $\bm{Y}$. The posterior probability $p_{m,n}$ defined in Eq.  \eqref{eq:pmn} indicates the coincidence between $\bm{y}_m$ and $\bm{x}_n$. For the probabilities in a column where $\bm{x}_n$ is fixed, they follow that $\sum_{m=1}^{M}{p_{m,n}}=1$. Ideally, the probability $p_{m^*,n}=1$ proves that $x_n$ only corresponds to the $y_{m^*}$. In the case depicted in Figure \ref{fig:probabilitymatrix_a}, this narrowest diagonal band shows a one-to-one correspondence between $\bm{X}$ and $\bm{Y}$, and the probabilities inside and outside the diagonal band have the greatest difference.

In the following, we design three different distance measures based on geometry and statistics, respectively.

\subsubsection{Low-rank measure}
\label{sec:lowrank}
The measure attempts to characterize the diagonal band. We strip the diagonal band from the probability matrix $\bm{P}$, keeping or eliminating the diagonal band. A significantly effective elimination method is that the higher probabilities are considered as noise in $\bm{P}$, and the low-rank matrix does not retain this information, widely applied in image denoising. 

Robust principal component analysis (RPCA) \cite{wright2009robust}, as a low-rank representation solver, performs a vital function in compressive sensing and sparse representation against traditional methods such as PCA. The key is the decomposition of the complex matrix $\bm{C}$ into a low-rank matrix $\bm{A}$ and a sufficiently sparse error matrix $\bm{E}$ shown in Eq. \eqref{eq:rpca}.
\begin{equation}
\label{eq:rpca}
    \min_{\bm{A},\bm{E}}||\bm{A}||_* + \lambda||\bm{E}||_1, \ \ \ \ subject\ to\ \bm{C} = \bm{A} + \bm{E},
\end{equation}
where $||\cdot||_*$ denotes the nuclear norm, $||\cdot||_1$ denotes the sum of the absolute value of matrix elements, and $\lambda$ is a weighting parameter. There are various open-source RPCA variants and their evaluations\footnote{The RPCA variant in our work is maintained by the Perception and Decision Lab at the University of Illinois at Urbana-Champaign and Microsoft Research Asia in Beijing. You can find the methods and brief introductions we use at the following website : https://people.eecs.berkeley.edu/\~{}yima/matrix-rank/sample\_code.html}. Note that Lin et al. \cite{lin2010augmented} proposed a variant of RPCA based on inexact augmented Lagrange multiplier (IALM) method  -- a trade-off between higher precision and less storage/time, which works best for our method in practice.

\begin{figure}[htbp]
%\vspace{-0.0cm}
\centering
\begin{minipage}[b]{0.32\linewidth}
{\label{b453(X) b453(Y)}\includegraphics[width=1\linewidth]{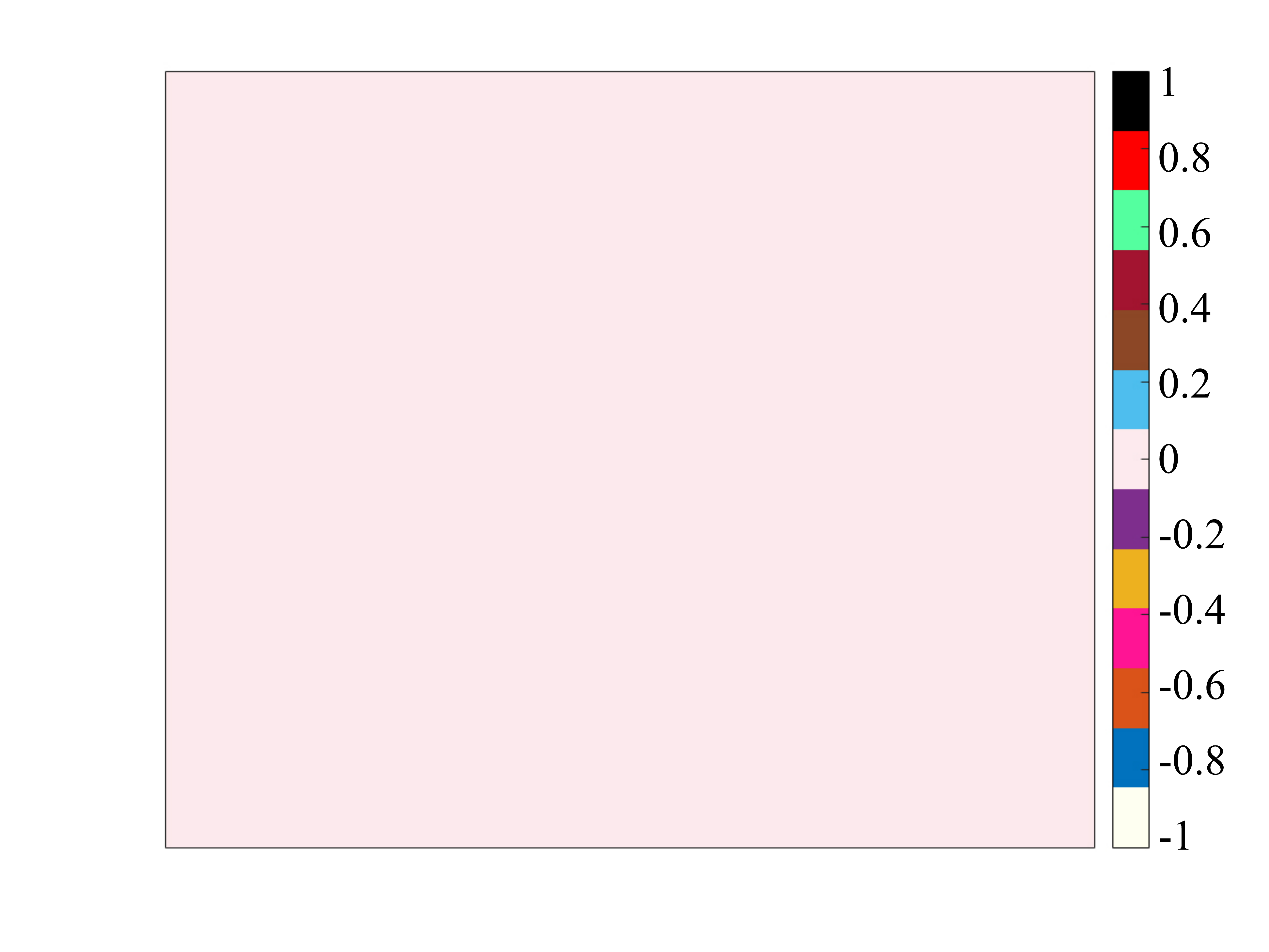}}
\end{minipage}
\begin{minipage}[b]{0.32\linewidth}
{\label{b227(X) b378(Y)}\includegraphics[width=1\linewidth]{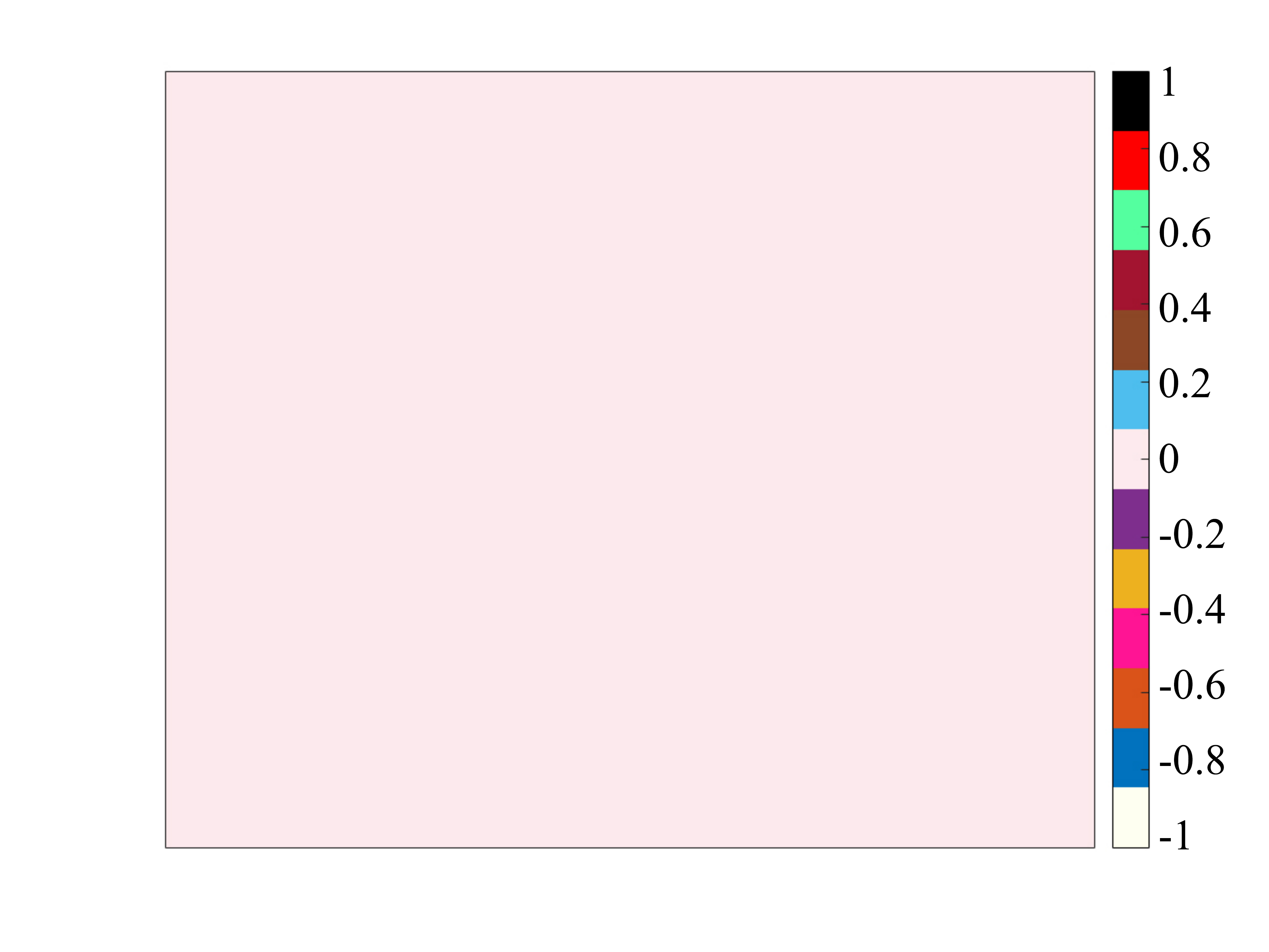}}
\end{minipage}
\begin{minipage}[b]{0.32\linewidth}
{\label{b453(X) d521(Y)}\includegraphics[width=1\linewidth]{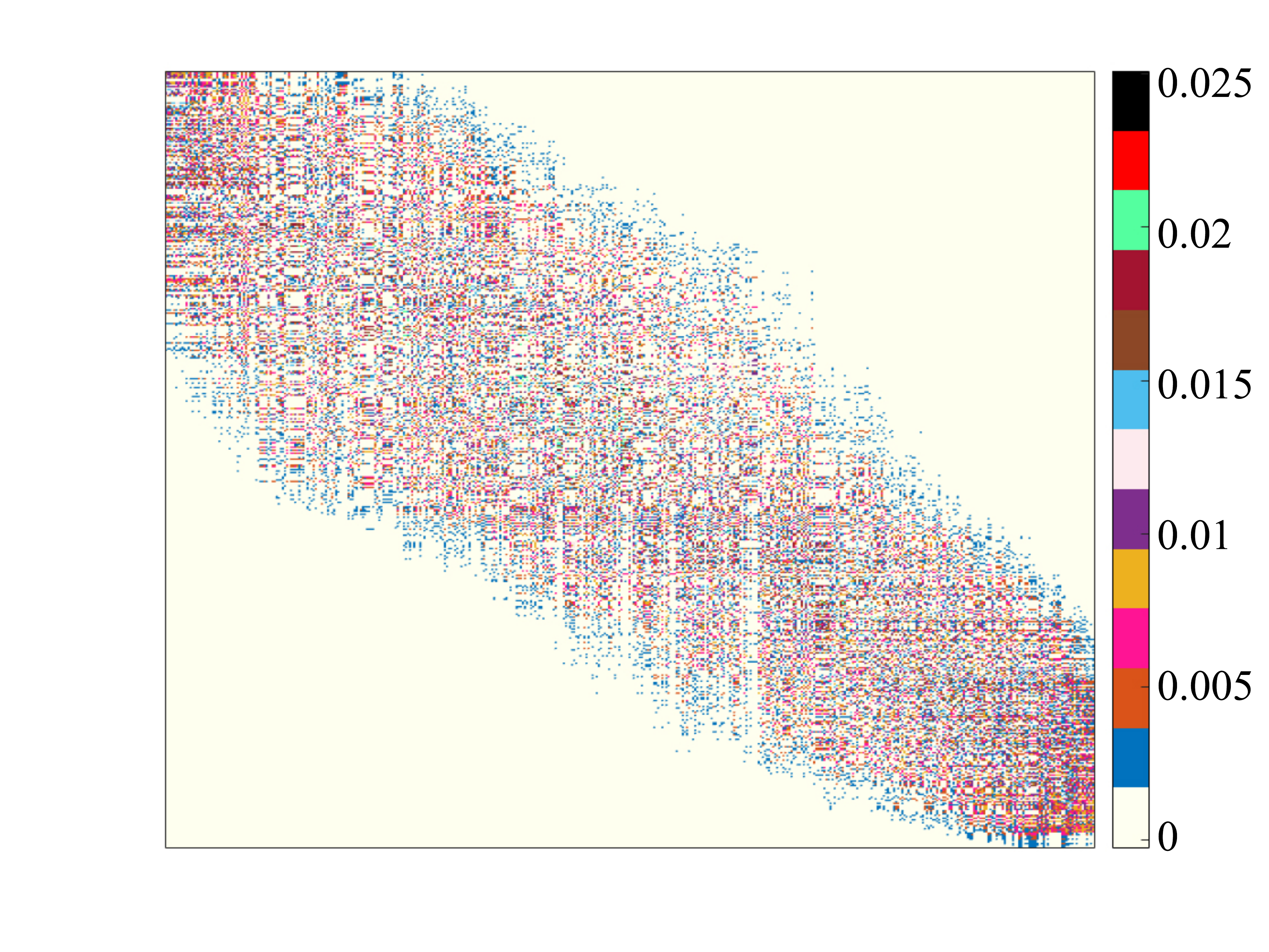}}
\end{minipage}
\caption{ The low rank matrices. For two almost consistent models, the low-rank matrix is even close to the all-zero matrix, which proves the role of the IALM method in model comparison. }
\label{fig:LR_PM}
\end{figure}

Figure \ref{fig:LR_PM} shows the low-rank components of the three probability matrices in Figure \ref{fig:probabilitymatrix}. For the complete or almost similar Bunny models, IALM characterizes the $\bm{P}$ as an all-zero matrix, while the low-rank matrix $\bm{A}$ corresponding to the registration of the Bunny model and the Dragon model retains most of the values.

To facilitate the measurement of the low-rank matrix, we define a distance named LR based on IALM in Eq. \eqref{eq:ID_LR}.

\begin{equation}
\label{eq:ID_LR}
    LR = \frac{1}{M}\frac{1}{N} \sum_{i=1}^{M}\sum_{j=1}^{N}[IALM(\bm{P}^T)]_{i,j}, 
\end{equation}
where $IALM(\cdot)$ denotes the function of obtaining low-rank matrix mentioned in Eq. \eqref{eq:rpca}, and ${(\cdot)}_{i,j}$ denotes the elements in the $i$-th row and the $j$-th column of a matrix.

\subsubsection{Kurtosis measure}
\label{sec:kurtosis}
This measure calculates the similarity between two point clouds in terms of the probability distribution of individual points. The colored bands in Figure \ref{fig:probabilitymatrix} indicate that $\bm{y_m}$ usually has a strong correlation with a short-range continuous $\bm{x_n}$.
Therefore, if each column of $\bm{P}$ is viewed as a data sample set, these values should follow certain patterns in the continuous space to some extent.

\begin{figure}[htbp]
%\vspace{-0.0cm}
\centering
\begin{minipage}[b]{0.325\linewidth}
\subfigure[b453(X) b453(Y)]{\label{fig:200column_a}\includegraphics[width=1\linewidth]{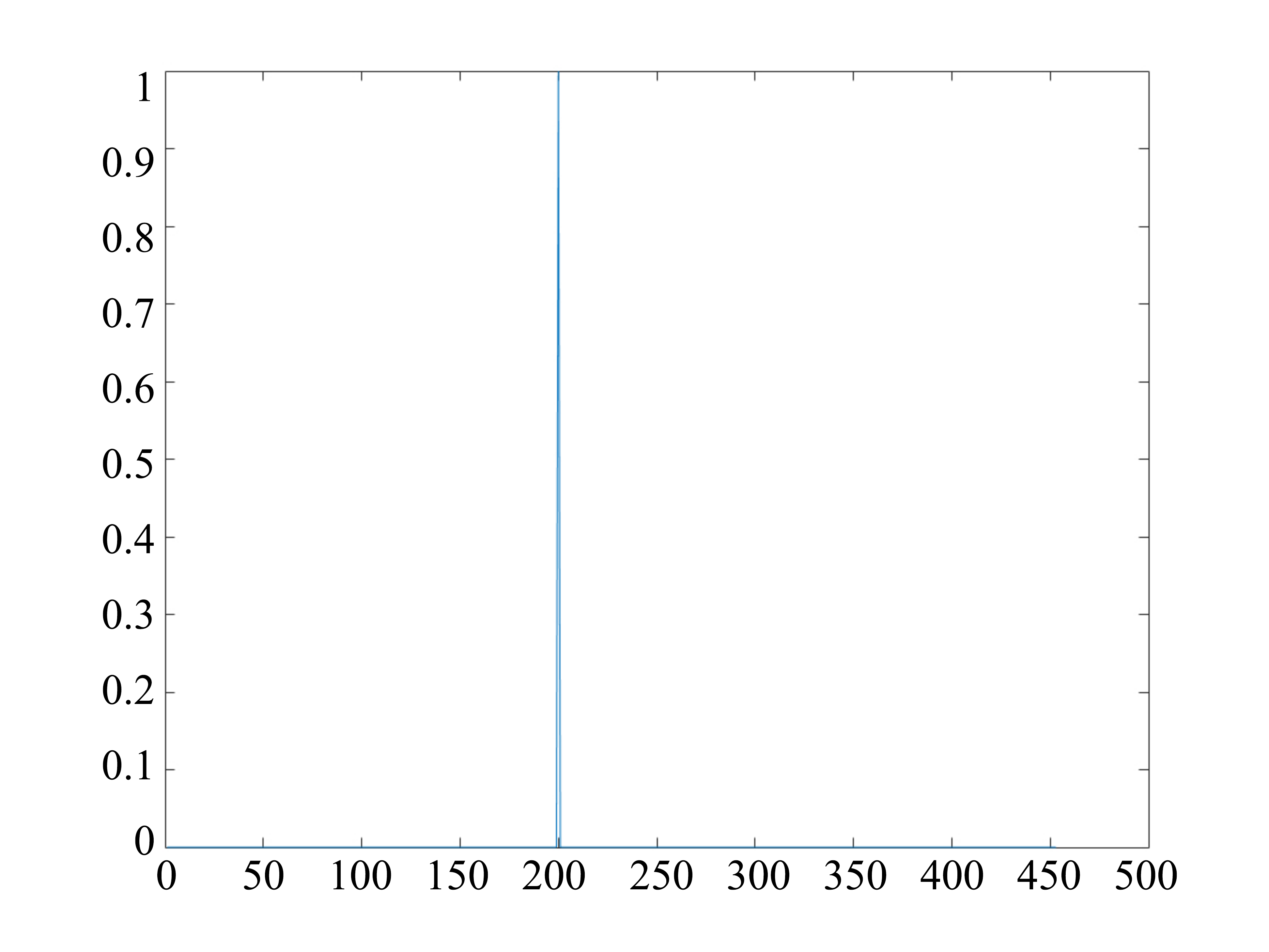}}
\end{minipage}
\begin{minipage}[b]{0.325\linewidth}
\subfigure[b227(X) b378(Y)]{\label{fig:200column_b}\includegraphics[width=1\linewidth]{figure/200b453xb453.pdf}}
\end{minipage}
\begin{minipage}[b]{0.325\linewidth}
\subfigure[b453(X) d521(Y)]{\label{fig:200column_c}\includegraphics[width=1\linewidth]{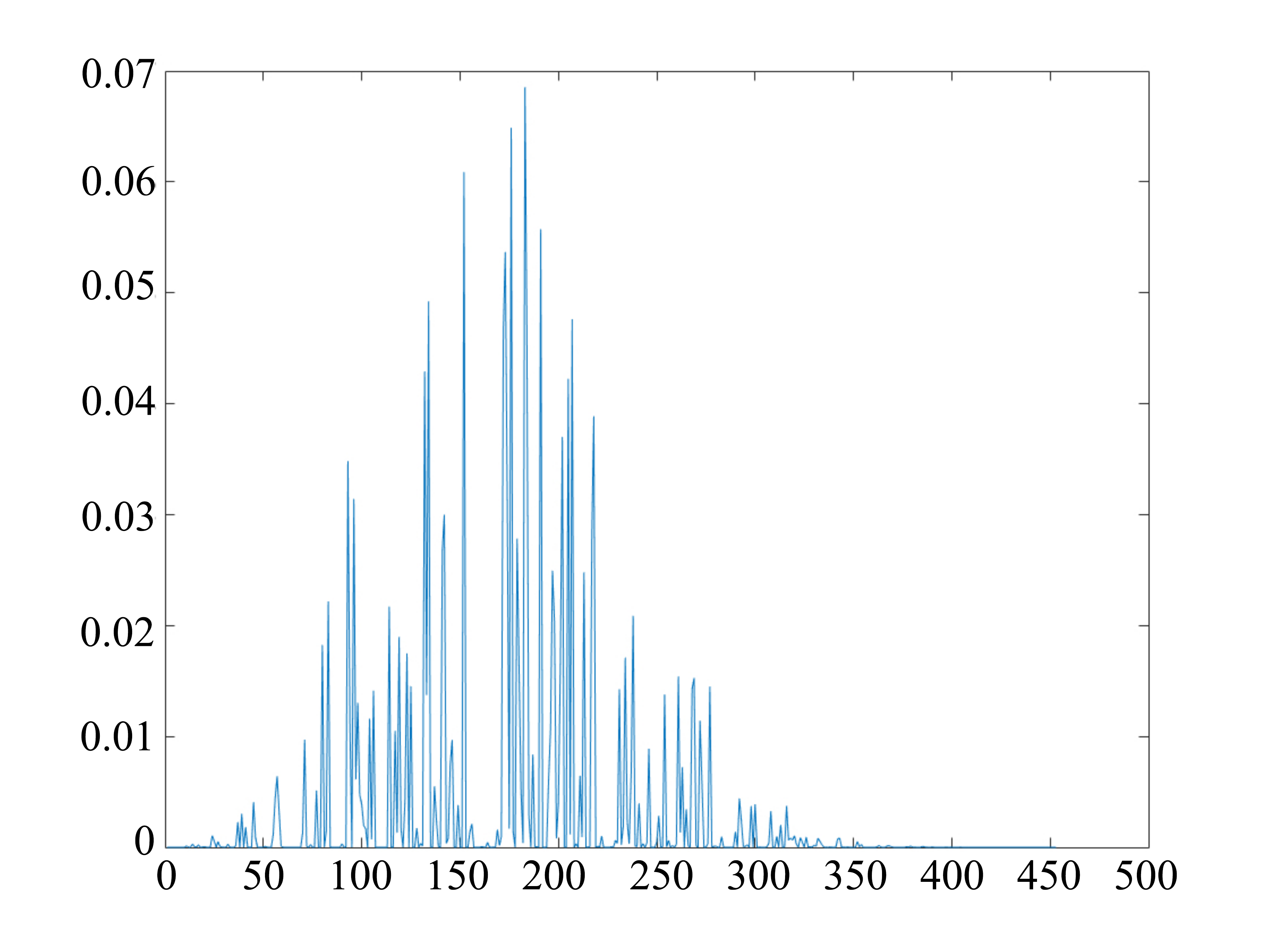}}
\end{minipage}
\caption{Distribution of the 200th column of the three probability matrices shown in Figure \ref{fig:probabilitymatrix}. 
}
\label{fig:200column}
%\vspace{-0.65cm}
\end{figure}

Figure \ref{fig:200column} describes the distribution of  $p(\bm{y}_m|\bm{x}_{200})$, where $\bm{y}_m$ can be any point in $\bm{Y}$.
And these distributions are generally low at both ends and high in the middle.
For the entirely aligned Bunny models shown in Figure \ref{fig:200column_a} and Figure \ref{fig:200column_b}, only $p_{200,200}=1$ while the rest elements are $0$.
Namely, closer pairs of points $(\bm{x_m},\bm{y_n})$ will induce larger probabilities, leading to a steeper distribution.
As for the registration of the Bunny model and the Dragon model, scattered outliers lead to a sharp decline in deviation (2 orders of magnitude).

We define the distance KURT based on the kurtosis in Eq. \eqref{eq:ID_Kurt}, which reveals and aggregates the sharpness of local distributions.
\begin{equation}
\label{eq:ID_Kurt}
KURT
= \frac{1}{N}\sum_{j=1}^N Kurt(\bm{P}_j),
\end{equation}
where $\bm{P}_j$ denotes the j-th column of $\bm{P}$, and $Kurt(\cdot)$ denotes the kurtosis coefficient of the given data calculated according to Eq. \eqref{eq:kurtosis}.
\begin{equation}
\label{eq:kurtosis}
    Kurt(\mathbf{x}) = \frac{\mu_4}{\sigma_4} = \frac{\frac{1}{M}\sum_{i=1}^M(x_i-\Bar{x})^4}
    {(\frac{1}{M}\sum_{i=1}^M(x_i-\Bar{x})^2)^2},
\end{equation}
where $\mu_4$ denotes the fourth central moment, and $\sigma$ denotes the standard deviation. 

\subsubsection{Correlation measure}
\label{sec:Correlation} 
This measure focuses on the correspondences of point clouds in geometric space. 
Correspondences is a collection of point pairs $(\bm{y}_m,\bm{x}_{n^*})$, which indicates $\bm{x}_{n^*}$ is most likely to $\bm{y}_m$ after point set registration. 
Consequently, $\bm{x}_{n^*}$ and $\bm{y}_m$ are observed as characteristics of $\bm{X}$ and $\bm{Y}$ in the geometric space, which is of great significance for judging the similarity of various models. In short, the correspondences reconstruct $\bm{X}$ in order to simulate $\bm{Y}$.

The process is as follows: a) generate matrices $X^{*}_{M \times 3}=\{\bm{x}_n*\}$ and $Y^{*}_{M \times 3}=\{\bm{y}_m\}$ respectively according to the correspondences, where the position of $\bm{x}_n*$ in $X^{*}$ is the same as the position of $\bm{y}_m$ in $Y^{*}$; 
b) uses the Pearson distance to measure the distance between $X^{*}$ and $Y^{*}$.
As a result, the irrelevance included in $\bm{X}$ and disturbance such as different numbers of points are eliminated.

We define the distance CORR based on Pearson distance in Eq. \eqref{eq:ID_Corr}.
\begin{equation}
\label{eq:ID_Corr}
    CORR = 
    \frac{\sum_{i=1}^M\sum_{j=1}^3 \delta_{X^*_{i,j}}\delta_{Y^*_{i,j}}}
    {\sqrt{\big[ \sum_{i=1}^M\sum_{j=1}^3 \delta_{X^*_{i,j}}^2\big]
    \big[\sum_{i=1}^M\sum_{j=1}^3 \delta_{Y^*_{i,j}}^2\big]}},
\end{equation}
where $\delta_{X^*_{i,j}} = X^*_{i,j} - \Bar{X}^*$, $\delta_{Y^*_{i,j}} = Y^*_{i,j} - \Bar{Y}^*$, ${(\cdot)}_{i,j}$ denotes the element in the $i$-th row and the $j$-th column of a matrix. $\Bar{X}^*$ and $\Bar{Y}^*$ denote the mean of all elements in $X^*$ and $Y^*$, respectively.

\textbf{\textit{Remark.}} Considering performance and needs, we design three different metrics to measure the similarity between two aligned point clouds. The LR distance is more stable in various attack scenarios, focusing on the essential information of the probability matrix. However, because of complex operations such as SVD operation, its required resources and running time will become unaffordable with the increase of data points. 
For this reason, we design the KURT distance and the CORR distance to calculate similarity quickly. Meanwhile, the CORR distance is naturally compact with the correlation which is the popular metric for evaluating the robustness of watermarking methods. It offers plausible comparisons with watermarking methods. 
However, the accuracy of the CORR distance is related to the number of points and alignment accuracy. When the number of points is insufficient, the CORR may fail to express the correlation accurately. Furthermore, two different models may still cause considerable overlap, which may induce a ``misleading'' CORR measure.

\subsection{Acceleration Strategies}
\label{sec:AccelerationStrategy}
The registration and RPCA involve expensive computation. As depicted in \cite{myronenko2010point}, the CPU time occupied by CPD increases exponentially as points soar. 
Meanwhile, the IALM method becomes unstable, and its performance drops sharply when the models are extraordinarily mismatched or have excessive points. Hence, we design speeding-up strategies in this section.

\subsubsection{Downsampling}
\label{sec:Downsampling}
As shown in Figure \ref{fig:overallflowchart}, point cloud after downsampling maintains original geometric characteristics but reduces the amount of data. This way alleviates the stress of calculation. 
Compared to random downsampling, the hierarchical expectation maximum (HEM) \cite{preiner2014continuous} algorithm tends to aggregate points level by level and provides a more meaningful representation for the original point cloud. The HEM algorithm is shown below. 
We utilize the Gaussian mixture model to characterize the 3D point clouds, applying only one initial EM iteration on the point cloud and then sequentially shrinking the mixture by hierarchically applying EM on Gaussians rather than points. 
Specifically, we pick up 1/3 of the Gaussian components obtained in the current round as the Gaussian components used in the next iteration, and the remaining 2/3 components are regarded as ``virtual samples''. Moreover, when executing the EM step, we only merge the Gaussian components in the same neighborhood.
In our implementation, the neighborhood radius is determined by the diagonal length of the minimum bounding box of the point cloud and a customized weight. 

\subsubsection{Segmentation and fusion}
\label{sec:Segmentation}
Another way to reduce the overhead caused by IALM is segmentation and fusion. Segmentation and fusion can effectively decrease the runtime. Generally, parts of two registered point clouds should be aligned respectively. With this premise, we design the segmentation and fusion strategy described as follows. First and foremost, we put two point clouds in the same three-dimensional coordinate system after registration and arrangement and then determine the longest axis from the x, y, and z-axis where the point clouds have the most extensive projection range. Later, the models are partitioned into $\max(\lceil M/T \rceil, \lceil N/T \rceil)$ parts evenly along the longest axis, in the sense that the number of points in each part is roughly $T$. Moreover, the last segment covers the remaining points to maintain the integrity of the point cloud. Finally, we calculate the similarity distance of each corresponding parts pair and merge their distances. This strategy simplifies the RPCA problem when processing relatively large point clouds.  
\section{Experimental Results}
\label{sec:results}
In this section, we illustrate the effectiveness of our method in point cloud data copy detection. We compare our method with 3D watermarking and 3D shape retrieval methods most relevant to this work. We adopt the attacks introduced by \cite{wang2010benchmark} in our experiments. 

\subsection{Comparison With 3D Model Watermarking}
\label{ComparisonWith3DModelWatermarking}
 
This section compares our method with the watermarking algorithm based on Euclidean distance deformation (EU) \cite{amar2016euclidean}. 

\jq{
Considering that this paper's core problem is copy detection, we only focus on the robustness of our method and the watermarking methods in detecting copies suffering from different attacks. }

\jq{It now boils down to the comparability of the watermark's correlation/BER and similarity distance. We determine the CORR distance threshold ($0.9951$) according to the principles of maximizing $TPR-FPR$, introduced in Section \ref{sssec:threshold}. Moreover, the CORR distance is greater than or equal to the threshold, indicating that the test model is a copy of the given model. Our method can hardly determine the threshold for the watermarking method because of its high sensitivity to attack intensity and the lack of statistics. However, it seems plausible that we use the false positive rate (FPR) of the CORR threshold to determine the threshold of the watermarking method. For a 64-bit watermark, if a misjudgment of $4\%$ (approximately the FPR of the CORR threshold) is allowed, the bit error rate (BER) should not be higher than $0.22$. }

\jq{Table \ref{table:watermarking_model} shows three models for this experiment. The homologous models are generated by the benchmark developed in \cite{wang2010benchmark}.}

\begin{table}[thbp]\tablefont
    \caption{Models used for 3D watermarking comparison. }
    \label{table:watermarking_model}
    \centering
    \begin{tabular}{c c c}
    \toprule
         Name & vertices & faces \\
    \midrule
        %  Bunny & 34,835 & 69,666\\
        %  Horse &  112,642 & 225,280\\
         Dragon & 50,000 & 100,000\\
         Cow & 2,904 & 5,804\\
         Hand & 36,619 & 72,958\\
    \bottomrule
    \end{tabular}
\end{table}

\begin{table}[thbp]\tablefont
    \centering
    \caption{Comparison of our method (CORR) with the EU approach (BER). }
    \label{table:wtermarkingEU}
    \scalebox{0.93}{
    \begin{tabular}{c c c c c c c c c}
    \toprule
    \multicolumn{2}{c}{Attack} & \multicolumn{2}{c}{Cow} & \multicolumn{2}{c}{Dragon} & \multicolumn{2}{c}{Hand}\\ \cmidrule(lr){1-2} \cmidrule(lr){3-4} \cmidrule(lr){5-6} \cmidrule(lr){7-8}
    Type & Intensity & EU & CORR & EU & CORR & EU & CORR\\
    \midrule
    \multirow{4}{*}{NA} 
    & 0.05\% & 0.08 & 1.0000 & 0 & 0.9999 & 0 & 1.0000\\
    & 0.1\%  & 0.16 & 1.0000 & 0.04 & 0.9999 & 0.01 & 1.0000\\
    & 0.3\%  & 0.23 & 1.0000 & 0.07 & 0.9999 & 0.27 & 1.0000\\
    & 0.5\%  & 0.31 & 1.0000 & 0.20 & 0.9999 & 0.26 & 1.0000\\
    \midrule
    \multirow{4}{*}{QU} 
    &10-bits & 0.02 & 1.0000 & 0.07 & 0.9999 & 0.10 & 1.0000\\
    &9-bits & 0.14 & 1.0000 & 0.16 & 0.9999 & 0.32 & 1.0000\\
    &8-bits  & 0.22 & 1.0000 & 0.20 & 0.9999 & 0.26 & 1.0000\\
    &7-bits  & 0.35 & 1.0000 & 0.26 & 0.9998 & 0.10 & 1.0000\\
    \midrule
    \multirow{4}{*}{SM} 
    & 5 & 0.10 & 0.9999 & 0.13 & 0.9999 & 0.01 & 1.0000\\
    & 10 & 0.16 & 0.9999 & 0.16 & 0.9999 & 0.10 & 1.0000\\
    & 30 & 0.23 & 0.9997 & 0.20 & 0.9999 & 0.20 & 1.0000\\
    & 50 & 0.36 & 0.9996 & 0.23 & 0.9999 & 0.23 & 1.0000\\
    \midrule
    \multirow{3}{*}{CR} 
    & 10\% & 0  & 0.9935 & 0 & 0.9930 & 0.01 & 0.9922\\
    & 30\%  & 0.16 & 0.9655 & 0.13 & 0.9452 & 0.10 & 0.9713\\
    & 50\%  & 0 & 0.9316 & 0.20 & 0.8684 & 0.13 & 0.7967\\
    \midrule
    \multirow{5}{*}{SI} 
    & 50.0\% & 0.16 & 0.9879 & 0.23 & 0.9995 & 0.20 & 0.9913\\
    & 70.0\%  & 0.22 & 0.9881 & 0.29 & 0.9993 & 0.16 & 0.9933 \\
    & 90.0\%  & 0.26 & 0.9875 & 0.46 & 0.9990 & 0.16 & 0.9985\\
    & 95.0\%  & 0.13 & 0.9796 & 0.53 & 0.9983 & 0.26 & 0.9953 \\
    & 97.5\%  & 0.20 & 0.9571 & 0.56 & 0.9981 & 0.30 & 0.9676\\
    \midrule
    \multirow{3}{*}{SU} 
    & Loop & 0.07 & 0.9999 & 0.07 & 0.9999 & 0.20 & 1.0000\\
    & Midpoint & 0 & 0.9999 & 0.08 & 0.9999 & 0 & 1.0000\\
    & $\sqrt{3}$ & 0.02 & 0.9999 & 0.23 & 0.9999 & 0.04 & 1.0000\\
    \bottomrule
    \end{tabular}}
\end{table}

\jq{Table \ref{table:wtermarkingEU} lists the comparisons between our method and the EU method against three geometric attacks, including noise, quantization and smoothing, and three connectivity attacks, including cropping, simplification and subdivision. Our method outperforms the EU method against geometric attacks, since CORR almost all reaches the maximum value of $1$. With the increase of attack intensity in the EU method, BER increases significantly and exceeds the threshold. 
With the intensifying of attack, BER rises considerably and exceeds the threshold. For example, the BER of the cow model with $0.5\%$ noise reaches 0.31. 
Regarding the subdivision attack, our method is also superior to the EU method. Although the BER of the EU method watermark does not exceed the above threshold of $0.22$ except for the dragon model suffering from the $\sqrt{3}$ subdivision, CORR almost perfectly reaches the maximum value ($1$). Furthermore, both methods have their own merits against simplification attacks. The EU method is more suitable for the cow and hand models, while our method resists all the simplified scenarios of the dragon model. As for cropping attacks, our method performs poorly. However, the results indicate that our method is more robust than the EU method in most cases. }

\subsection{Comparison with 3D Shape Retrieval}
\label{comparisonWith3DShapeRetrieval}
 
We compare MVCCN \cite{su2015multi}, MeshNet \cite{feng2019meshnet} and USC \cite{tombari2010unique} with our method. For fair comparisons, we design our own homologous model set HM25 based on the Princeton ModelNet40 \cite{wu20153d} and then evaluate these approaches.

\subsubsection{Homologous Model Dataset}
\begin{figure}[htbp]
%\vspace{-0.0cm}
\centering
\begin{minipage}[b]{0.3\linewidth}
\subfigure[origin]{\label{fig:attack_origin}\includegraphics[width=1\linewidth]{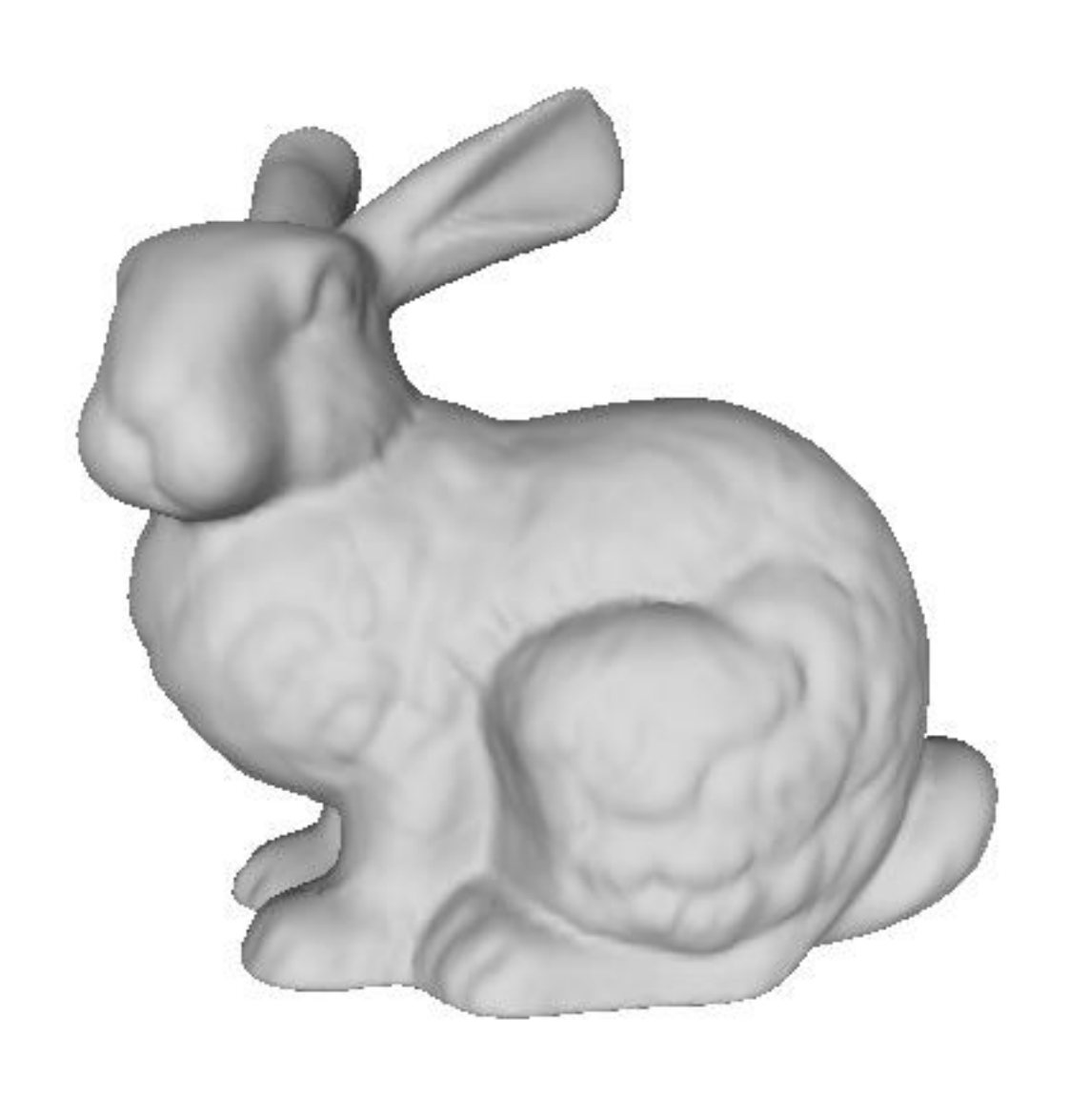}}
\end{minipage}
\begin{minipage}[b]{0.3\linewidth}
\subfigure[crop]{\label{fig:attack_crop}\includegraphics[width=1\linewidth]{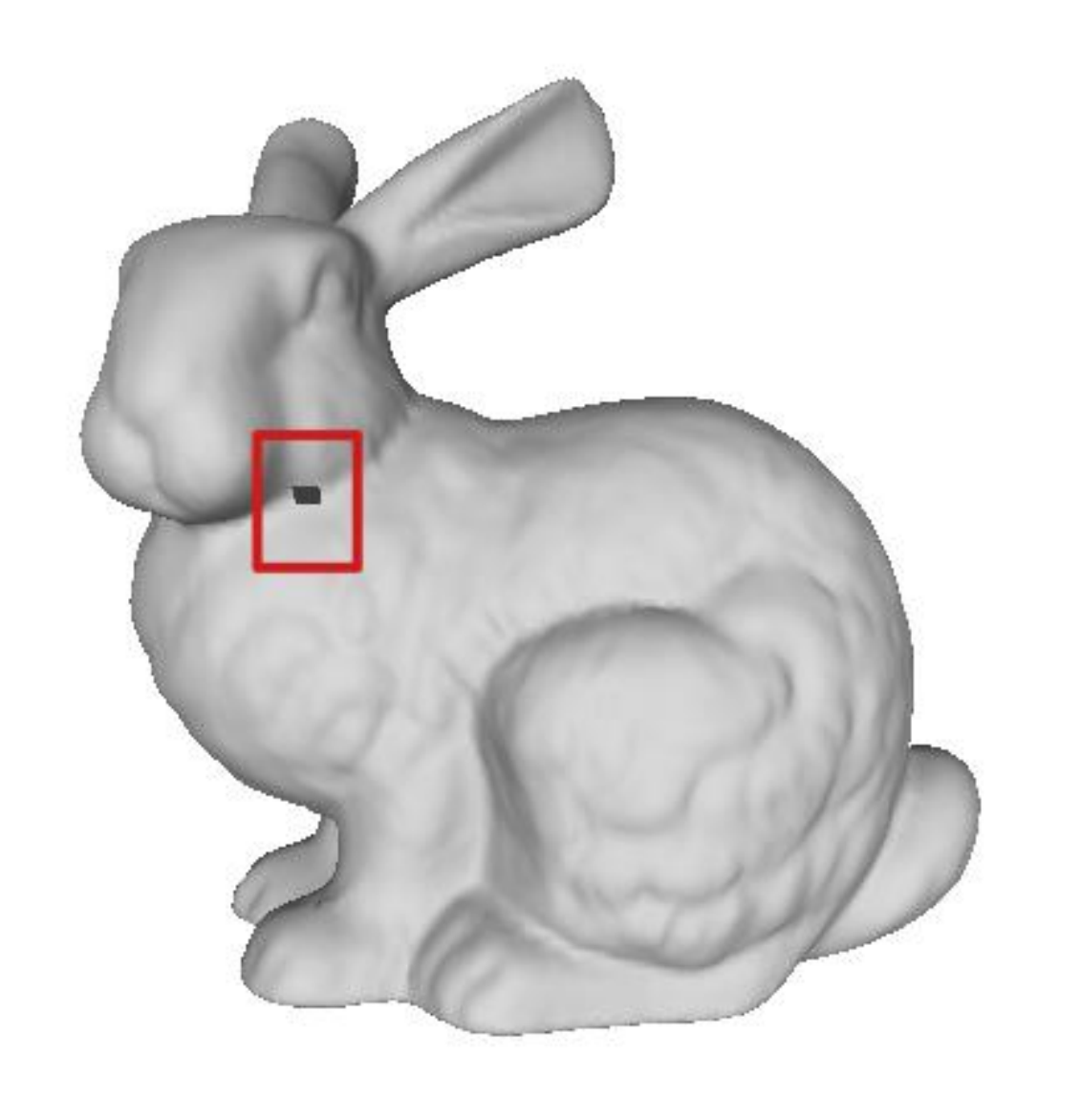}}
\end{minipage}
\begin{minipage}[b]{0.3\linewidth}
\subfigure[noise]{\label{fig:attack_noise}\includegraphics[width=1\linewidth]{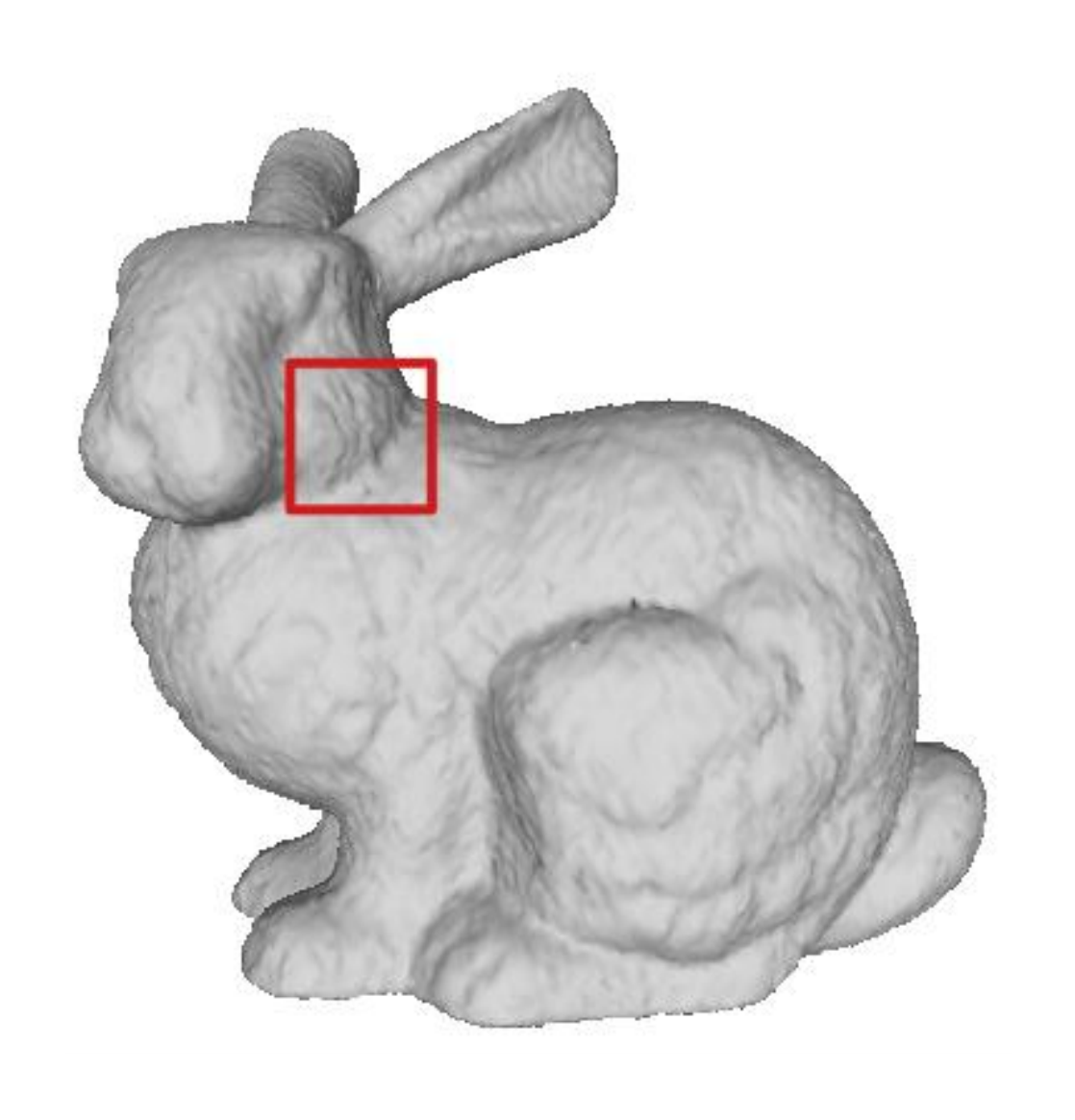}}
\end{minipage}
\begin{minipage}[b]{0.3\linewidth}
\subfigure[reorder]{\label{fig:attack_reorder}\includegraphics[width=1\linewidth]{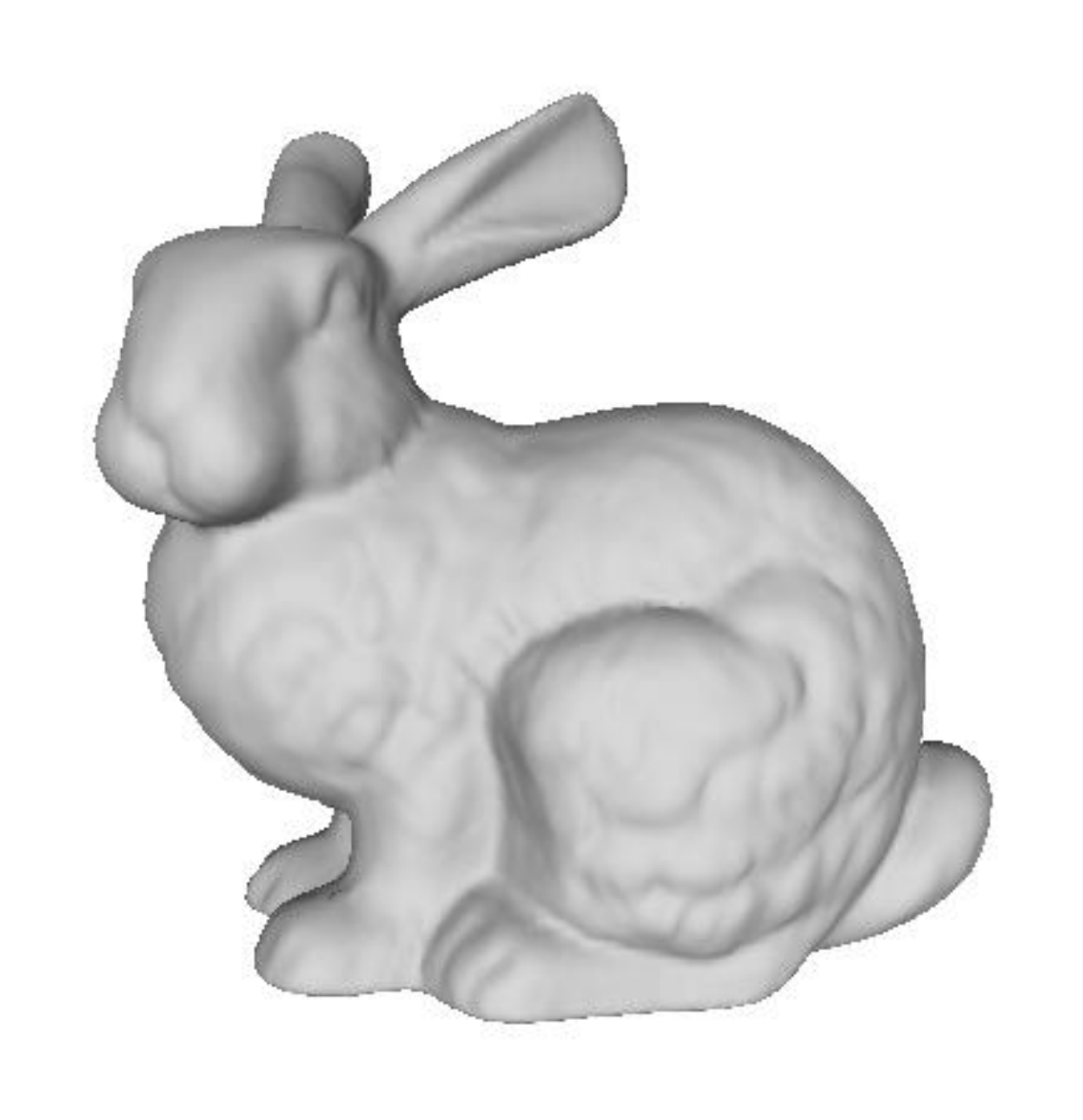}}
\end{minipage}
\begin{minipage}[b]{0.3\linewidth}
\subfigure[quantization]{\label{fig:attack_quan}\includegraphics[width=1\linewidth]{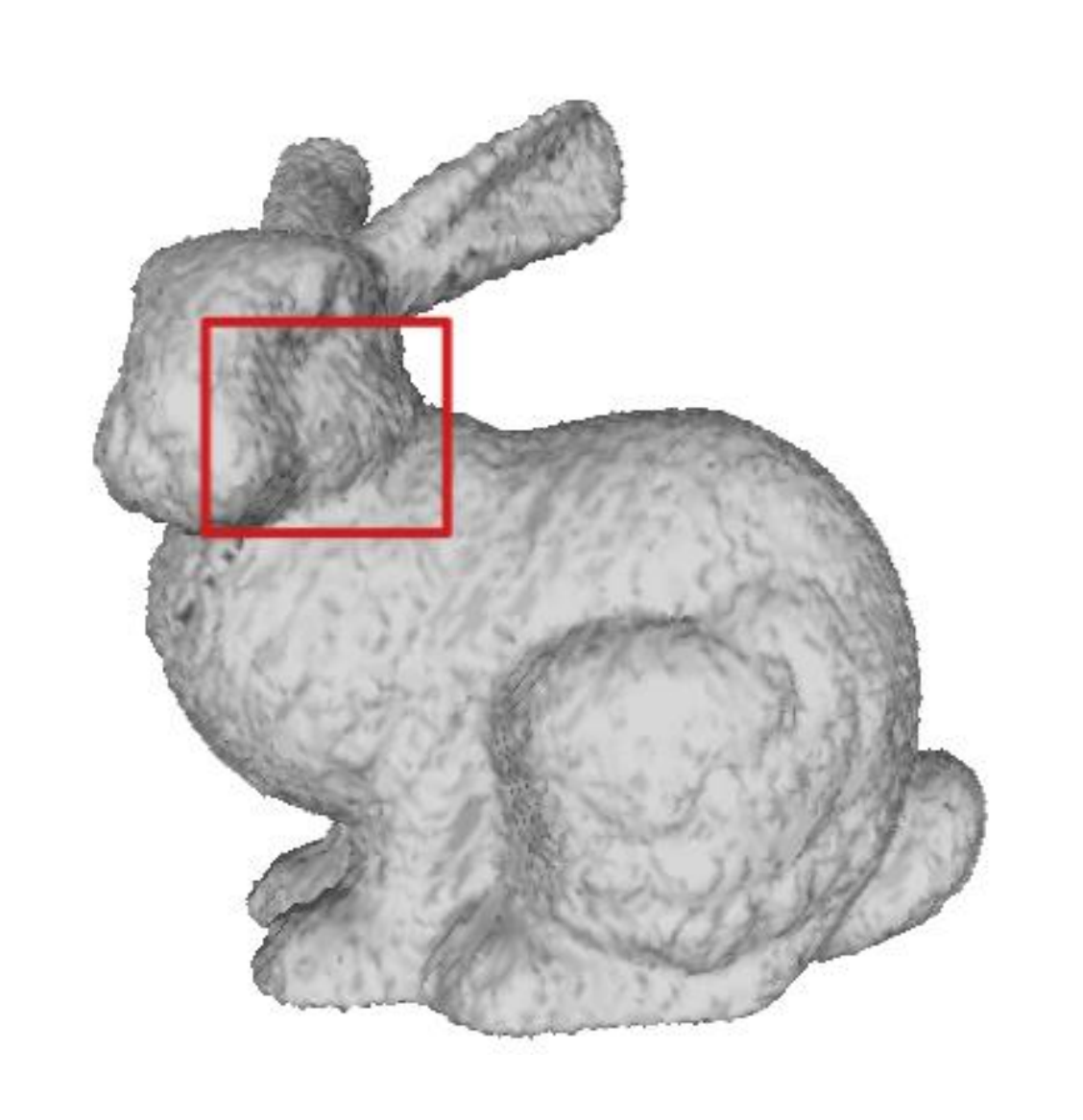}}
\end{minipage}
\begin{minipage}[b]{0.3\linewidth}
\subfigure[similarity]{\label{fig:attack_similarity}\includegraphics[width=1\linewidth]{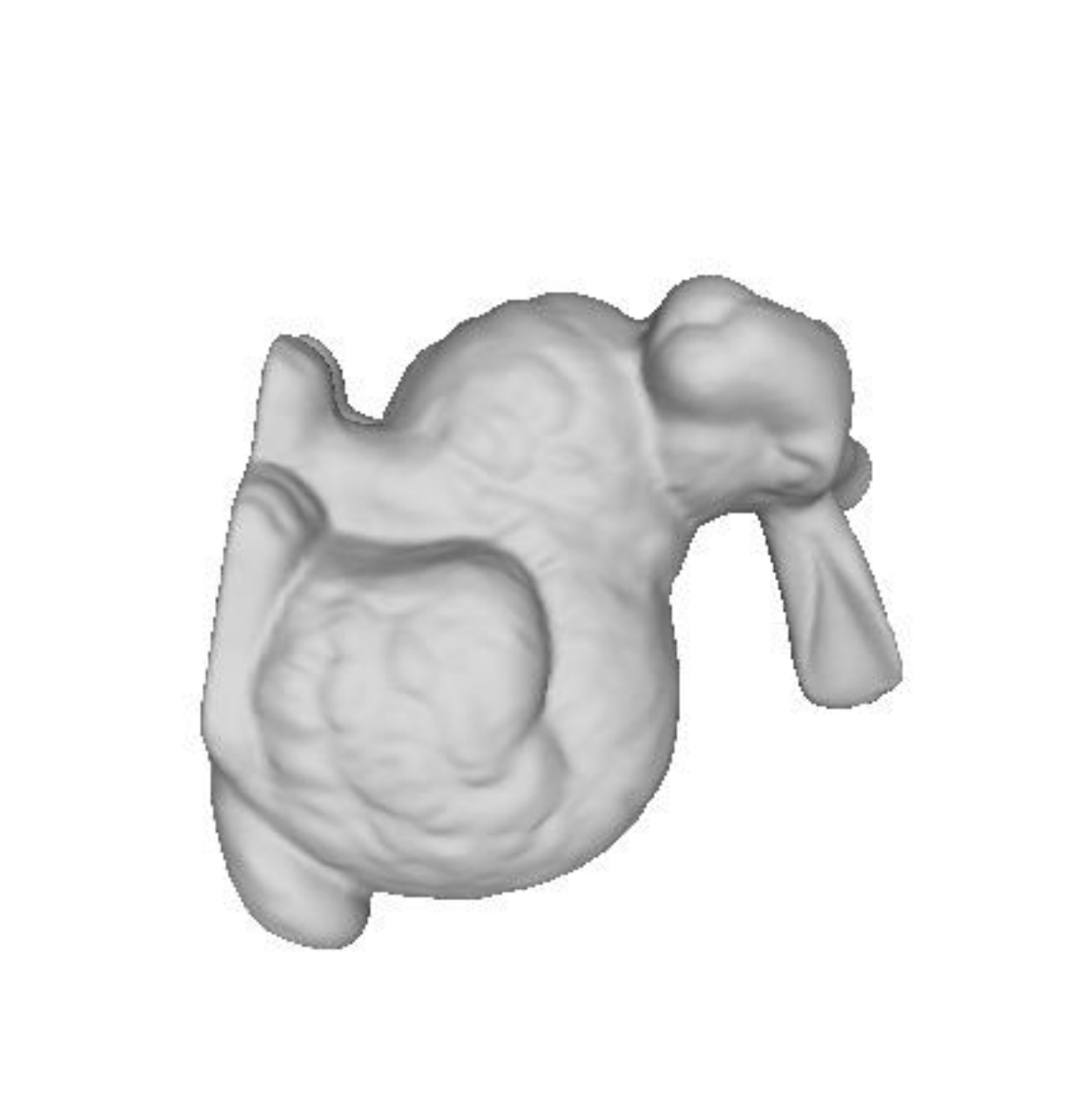}}
\end{minipage}
\begin{minipage}[b]{0.3\linewidth}
\subfigure[simplification]{\label{fig:attack_simplification}\includegraphics[width=1\linewidth]{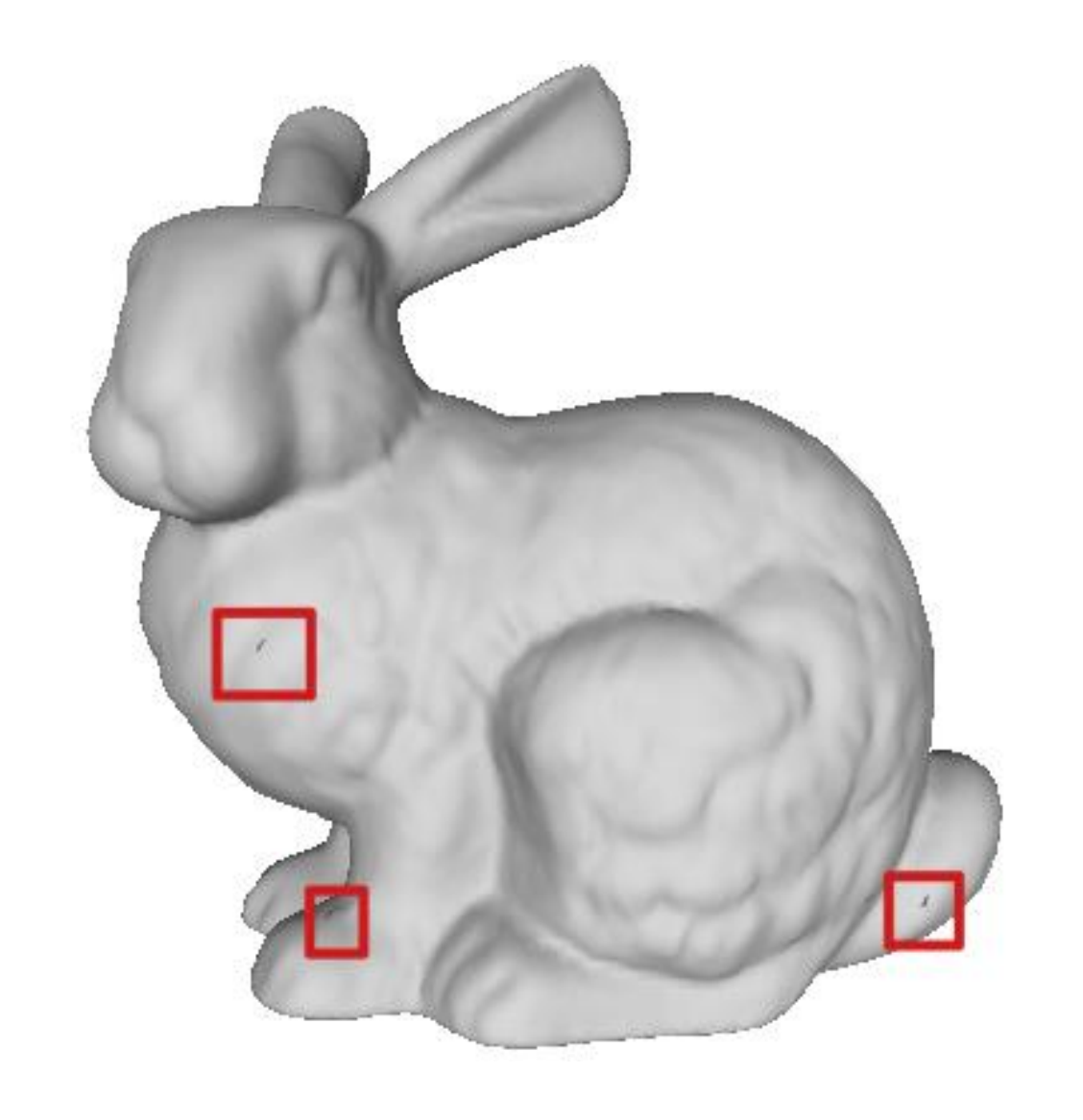}}
\end{minipage}
\begin{minipage}[b]{0.3\linewidth}
\subfigure[smooth]{\label{fig:attack_smooth}\includegraphics[width=1\linewidth]{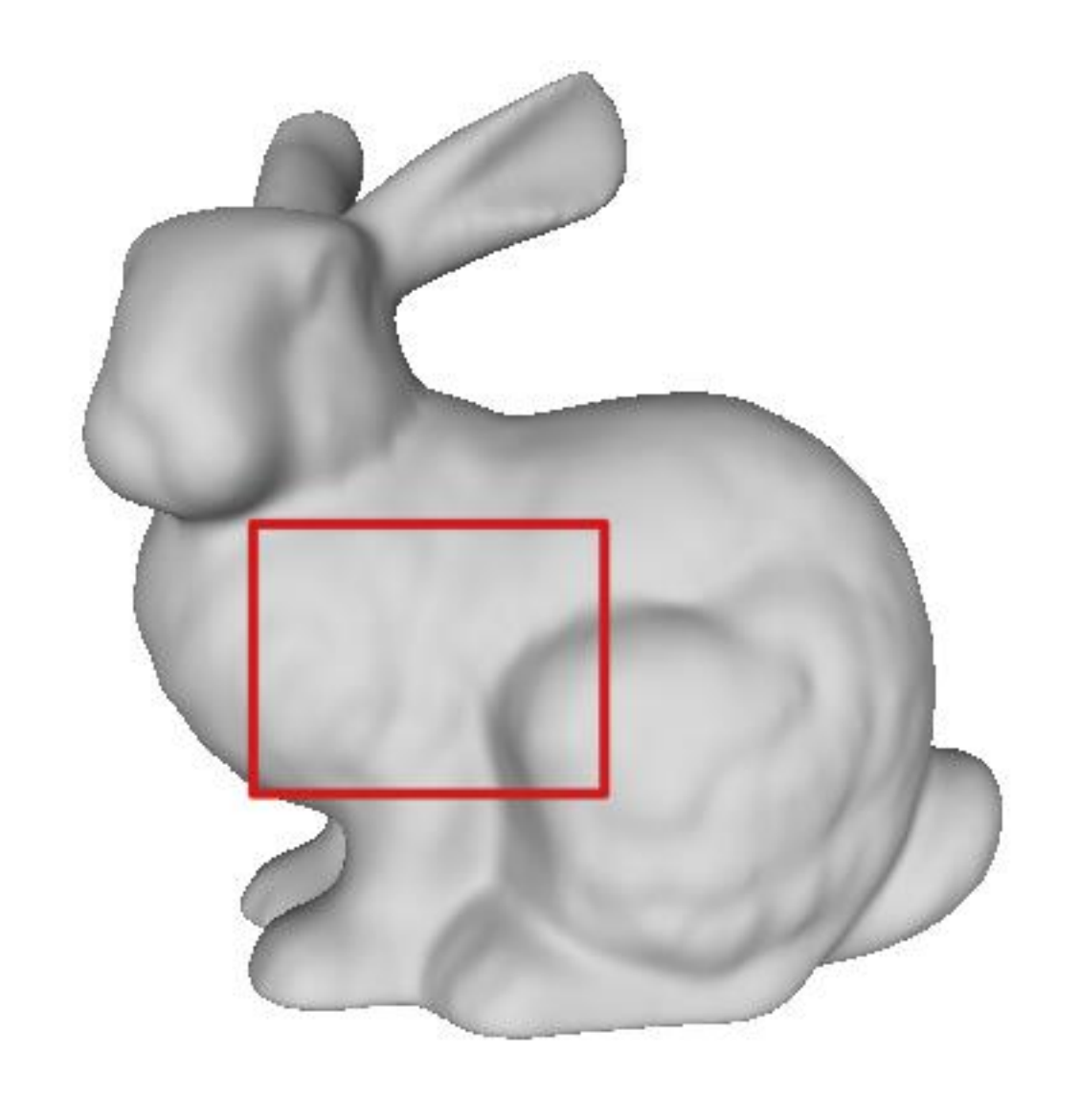}}
\end{minipage}
\begin{minipage}[b]{0.3\linewidth}
\subfigure[subdivision]{\label{fig:attack_subdivision}\includegraphics[width=1\linewidth]{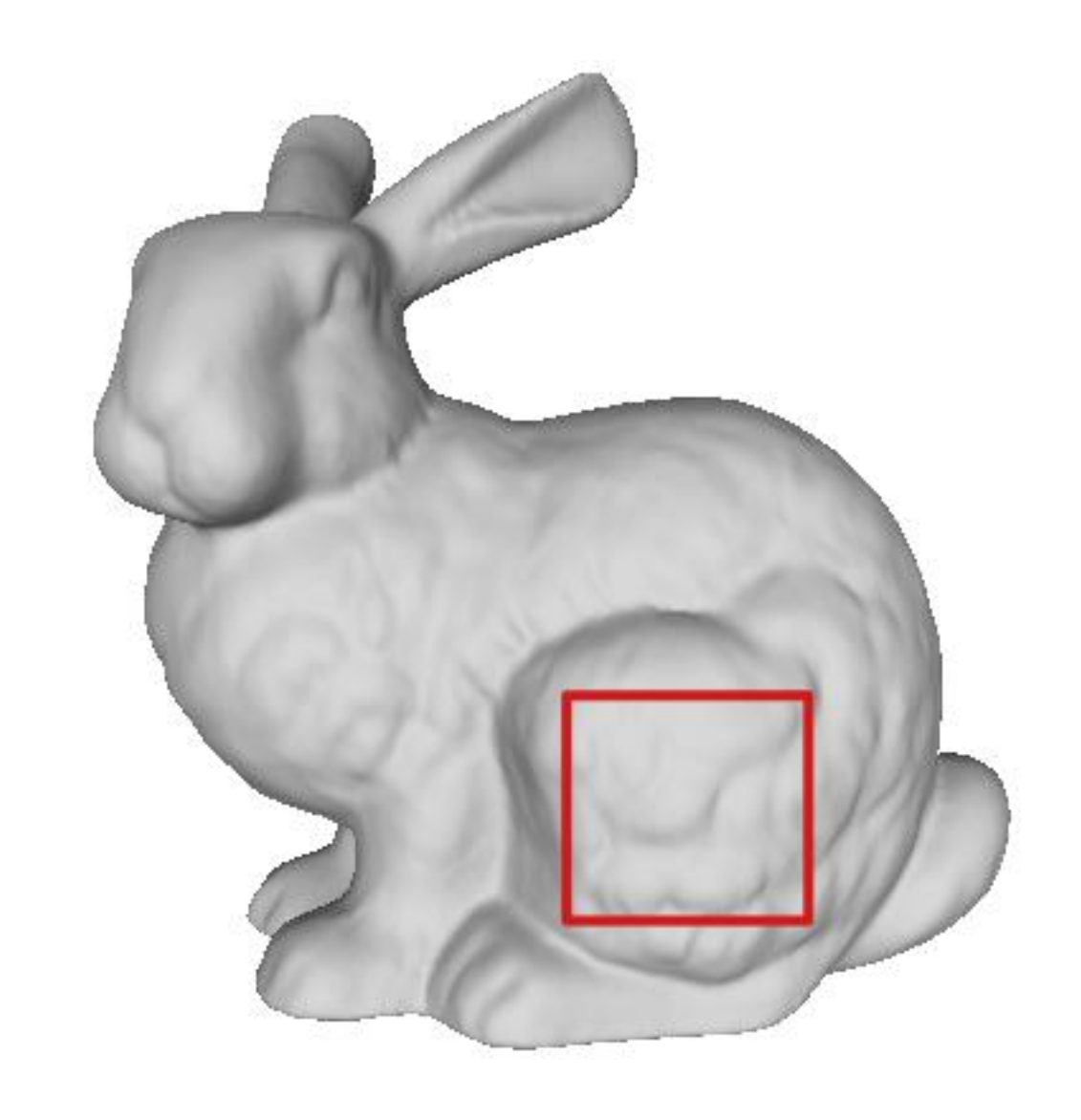}}
\end{minipage}
\caption{Examples of the Bunny's homologous models. The red box marks the difference between the currently attacked model and the original model. }
\label{fig:attack}
\end{figure}

Previously, researchers congregate 40 categories of routinely accessed objects online to construct the ModelNet40 containing 12,311 shapes. However, we construct the HM25, which focuses on the above attacks rather than categories.   
We yield some homologous models with the attacks in \cite{wang2010benchmark}. And the homologous models are obtained by imposing diverse attacks on the same original model. Table \ref{table:Homoisomer} lists the attacks and parameters. And Figure \ref{fig:attack} shows eight attacked models of the Bunny model (i.e., Bunny's homologous models).
\begin{table}[thbp]\tablefont
    \centering
    \caption{Type and intensity of attacks on homologous models. Intensity implies various units/properties perchance. } 
    \label{table:Homoisomer}
    \begin{tabular}{l c c c}
    \toprule
     Attack Type & Abbr. & Intensity & Number \\ 
    \midrule
     Crop & CR & $5\%$/$10\%$ & $1$  \\
     Noise Addition & NA & $0.1\%$/$0.3\%$/$0.5\%$ & $1$  \\
     Quantization & QU & $9$/$8$/$7$ & $1$  \\
     Reorder & RE & / & 1  \\
     Simplification & SI & $10\%$/$20\%$/$40\%$ & 1  \\
     Smooth & SM & $10$/$30$/$50$ & 1 \\
     Similarity Transformation & ST & / &  3 \\
     Subdivision & SU & loop/$\sqrt{3}$/midpoint &  1 \\
    \bottomrule
    \end{tabular}
\end{table}

We pick up 625 objects in 25 categories from ModelNet40 as the source models and obtain 21 attacked models from each source model. Each category takes 25 source models and is split by a training-test ratio of 8:2.

It should be noted that this dataset generation procedure should comply with several principles as follows.  
a) The source training and test models are from the train set and test set of ModelNet40, respectively; 
b) All models, including homologous models, must not exceed 4096 faces considering MeshNet, which aggregates features of all faces; 
c) The attacks are only applicable to two-dimensional orientable manifold meshes which are not guaranteed by ModelNet40.  
At last, we created the HM25, containing 13,750 3D models of 25 categories.

\begin{table*}[thbp]\tablefont
    \centering
    \caption{Comparison of our method with other three retrieval methods. The first four categories are geometric attacks (ST, NA, SM, QU), and the latter two are connectivity attacks (CR, SI). NA1 denotes the $0.1\%$ noise attack (the rest of the symbols and so on). See Table \ref{table:Homoisomer} for abbreviations and Intensities. Bold indicates the highest performance. } 
    \label{table:retrieval}
    \scalebox{0.91}{
    \begin{tabular}{ c  c c c c c c  c c c c c c  c c c c c c }
    \toprule
    \multirow{2}{*}{Attack}  & \multicolumn{6}{c}{$Top_1$} & \multicolumn{6}{c}{$Top_2$} & \multicolumn{6}{c}{$Top_5$} \\ \cmidrule(lr){2-7} \cmidrule(lr){8-13} \cmidrule(lr){14-19} 
    & LR & Kurt & Corr & \cite{feng2019meshnet} & \cite{su2015multi} & \cite{tombari2010unique} & 
    LR & Kurt & Corr & \cite{feng2019meshnet} & \cite{su2015multi} & \cite{tombari2010unique} &
    LR & Kurt & Corr & \cite{feng2019meshnet} & \cite{su2015multi} & \cite{tombari2010unique}  \\ 
    \midrule
    ST 
    & 43.2 & 56.0 & $\mathbf{60.8}$ & 28.8 & 26.4 & 20.0 
    & 50.4 & $\mathbf{64.0}$ & 60.8 & 46.4 & 32.8 & 20.8 
    & 56.0 & 67.2 & $\mathbf{68.0}$ & 59.2 & 39.2 & 24.8 \\
    \midrule
    NA1 & 55.2 & 83.2 & 94.4 & 99.2 & $\mathbf{100}$ & 94.4 & 
    60.8 & 94.4 & 95.2 & $\mathbf{100}$ & $\mathbf{100}$ & 94.4 & 63.2 & 96.0 & 99.2 & $\mathbf{100}$ & $\mathbf{100}$ & 94.4  \\
    \midrule
    SM10 & 60.0 & 67.2 & 76.8 & 94.3 & $\mathbf{97.6}$ & 87.2 & 
    67.2 & 77.6 & 80.0 & 96.8 & $\mathbf{99.2}$ & 88.0 & 
    70.4 & 87.2 & 93.6 & $\mathbf{100}$ & $\mathbf{100}$ & 89.6 \\
    \midrule
    QU7 & 84.8 & 78.4 & 93.6 & 93.6 & $\mathbf{99.2}$ & 84.0 & 
    89.6 & 90.4 & 95.2 & 98.4 & $\mathbf{100}$ & 84.8 & 
    92.0 & 95.2 & 99.2 & $\mathbf{100}$ & $\mathbf{100}$ & 85.6 \\
    \midrule
    CR5 & 68.0 & 46.4 & 36.0 & 87.2 & $\mathbf{92.8}$ & 62.4 & 
    81.6 & 60.0 & 51.2 & 92.0 & $\mathbf{95.2}$ & 65.6 & 
    88.0 & 74.4 & 94.4 & 98.4 & $\mathbf{97.6}$ & 71.2 \\
    \midrule
    SI10 & 54.4 & 75.2 & 80.0 & 90.4 & $\mathbf{94.4}$ & 89.6 & 
    61.6 & 85.6 & 84.8 & 92.0 & $\mathbf{95.2}$ & 90.4 & 
    65.6 & 94.4 & 93.6 & $\mathbf{96.8}$ & 96.0 & 91.2 \\
    
    \midrule
    RE & 59.2 & 82.4 & 94.4 & 92.0 & $\mathbf{100}$ & 94.4& 
    64.8 & 94.4 & 95.2 & 98.4 & $\mathbf{100}$ & 94.4 & 
    66.4 & 96.0 & 99.2 & $\mathbf{100}$ & $\mathbf{100}$ & 94.4 \\
    \bottomrule
    % \hline
    \end{tabular}}
\end{table*}

\begin{figure*}
\center
\includegraphics[width=0.85\linewidth]{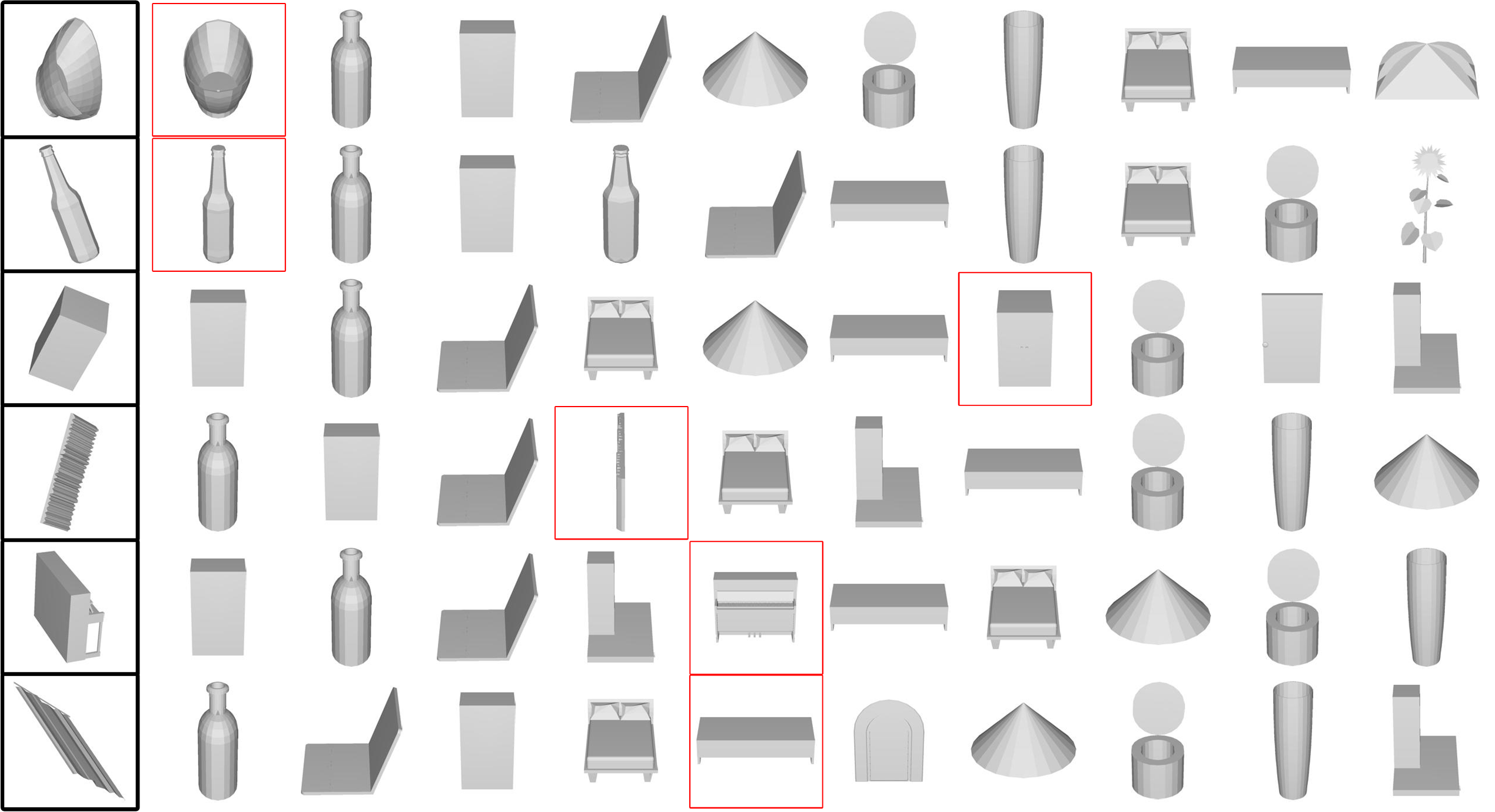}
\caption{Top ten original models queried by the homologous models based on the similarity transformation attack. The similarity is calculated by the CORR measure. 
%Retrieval results of similarity transform. 
And the black box is the query for similarity transformation and the red box indicates the best matched original model. %What's more, these models are arranged in order. 
}
\label{fig:retrievalresult}
\end{figure*}

\subsubsection{Experimental Setup}
We utilize HM25 to fine-tune MVCNN with 12 views pre-trained on ImageNet and  MeshNet pre-trained on the simplified version of ModelNet40 developed in \cite{feng2019meshnet}. The parameters are consistent with the original papers. We respectively extract the output of relu$_7$ layer of CNN$_2$ in MVCNN and the penultimate multilayer perceptron (MLP) in MeshNet as the descriptors of the 3D models. We then use the $L_2$ distance to evaluate similarity of different models. 

As for USC, instead of comparing the minimum Euclidean distance of point-wise descriptors, we implement USC with Point Cloud Library (PCL) \cite{rusu20113d} and employ a component analogous to maxpool to fuse the descriptors of 3D models. We also set the radius of USC based on the mean of the $20$-th nearest neighbor distance in the point cloud. Table \ref{table:descriptor} shows the descriptors of the three methods.

\begin{table}[thbp]\tablefont
    \centering
    \caption{Information for the compared methods. } 
    \label{table:descriptor}
    \begin{tabular}{l c c c}
    \toprule
    Methods & \tabincell{l}{
    Descriptor} & \tabincell{l}{
    Size} & \tabincell{l}{
    Measurement} \\ 
    \midrule
    MVCNN,12$\times$ & Relu7 & 86528 & Euclidean Distance
    \\
    MeshNet & Concat\_MLP & 1024 & Euclidean Distance
    \\
    USC & Maxpool & 1960 & Euclidean Distance
    \\
    \bottomrule
    \end{tabular}
\end{table}

In the experiment, we evaluate the performance of our method as well as 3D shape retrieval algorithms. Firstly, we divide the HM25 test dataset by attack types, leading to 21 query sub-datasets and a target sub-dataset ($\mathcal{T}$) containing the whole source models. Each sub-dataset contains 125 models. 
After that, we calculate the retrieval rate $Top_K$ of a given query database $\mathcal{Q}=\{q_1,\cdots,q_S\}$ on the target database $\mathcal{T}$. For the $i$-th model in $\mathcal{Q}$, we calculate the similarity between $q_i$ and the model in the $\mathcal{T}$ one by one (our method uses LR, KURT, and CORR distances, and other methods use the $L_2$ distance), and arrange the models in $\mathcal{T}$ in the descending order of similarity. The set of $K$ models most similar to $q_i$ is denoted as $r_i$. And $Top_K$ is defined as:
\begin{equation}
\begin{aligned}
\label{eq:TopK}
    Top_K &= \frac{1}{S}\sum_{i=1}^{S}\sum_{k=1}^{K}R_{i,k}\\
    R_{i,k} &= \begin{cases}
    1, & q_i \in \mathcal{H}(r_{i,k})\\
    0, & q_i \notin \mathcal{H}(r_{i,k})
    \end{cases},
\end{aligned}
\end{equation} 
where $S$ denotes the number of models in $\mathcal{Q}$ ($S=125$), $r_{i,k}$ is the $k$-th model in $r_i$, and $\mathcal{H}(x)$ denotes the collection of homologous models of the source model $x$.

\subsubsection{Result}
Table \ref{table:retrieval} summarizes the experimental results of shape retrieval on HM25. 
Our method outperforms other methods in combating similarity transformation. Compared with the USC method ($20.0\%$), the $Top_1$ retrieval rate of our method ($60.8\%$) is even three times higher than it. Besides, our method is robust against noise addition, quantization and reorder attacks, where $Top_5$ results of CORR measure reach above $99\%$, although somewhat inferior to the deep learning methods. However, there is a gap between our method and deep learning methods when facing connectivity attacks that adjoin or slit some points or edges. As for the cropping attack, the $Top_1$ accuracy of our method based on  CORR measure is as low as $36.0\%$ while MVCNN achieves $92.8\%$ accuracy. However, the CORR measure is up to $94\%$ in the $Top_5$ results against the connectivity attack. In short, the results demonstrate that our method performs remarkably in general, better than USC, and slightly poorer than MVCNN and MeshNet, in particular, $Top_2$ and $Top_5$ results. 

Regarding the performance of our method in Table \ref{table:retrieval}, we have the following observation. 
Our method may be weaker than MVCNN and MeshNet, which certainly benefit from the powerful deep learning. 
However, we must point out that the retrieval task is to search models in the same category of the given query, and it is challenging for the deep learning methods in the question of whether two 3D models are similar (e.g., partly similar, or the same model) regardless of the categories.   

To further illustrate the results intuitively, Figure \ref{fig:retrievalresult} presents some of the retrieval results of the queries based on similarity transformation according to the CORR measure. As shown in the third row of Figure \ref{fig:retrievalresult}, compared with the wardrobe and its original model, the wardrobe is more similar to the bottle. Several factors contribute to this unreasonable phenomenon. First, the wardrobe lacks enough points to describe its shape, and the CORR measure becomes invalid actually. Second, the CPD algorithm itself is not working well for the alignment.

\begin{table*}[thbp] \tablefont
    \centering
    \caption{Evaluation of downsampling strategy on our method. } 
    \label{table:Downsampling}
    \begin{tabular}{ccccccccc}
    \toprule
    \multicolumn{2}{c}{parameters} & \multicolumn{3}{c}{random} & \multicolumn{3}{c}{HEM} & benchmark\\
    \midrule
    \multirow{5}{*}{$\mathbf{X}$} & size & $13073\times 3$ & $2263\times3$ & $97\times3$ & $13073\times 3$ & $2263\times3$ & $97\times3$ & $34835\times3$ \\ %\cmidrule{2-9}
     & sampling rate($\%$) & 37.528 & 6.496 & 0.278& / & / & / & / \\ % \cmidrule{2-9}
     & layers & / & / & / & 3 & 3 & 10 & / \\ %\cmidrule{2-9}
     & amendatory factor & / & / & / & 2 & 4 & 10 & / \\ %\cmidrule{2-9}
    \midrule
     \multirow{5}{*}{$\mathbf{Y}$} & size & $15283\times 3$ & $2931\times3$& $115\times3$ & $15283\times 3$ & $2931\times3$& $115\times3$ & $34835\times3$ \\ %\cmidrule{2-9}
     & sampling rate($\%$) & 43.873 & 8.414 & 0.330 & / & / & / & / \\ %\cmidrule{2-9}
     & layers & / & / & / & 3 & 3 & 10 & /\\ %\cmidrule{2-9}
     & amendatory factor & / & / & / & 2 & 4 & 10 & /\\ %\cmidrule{2-9}
    \midrule
    \multicolumn{2}{c}{time of CPD(s)} & 336.35 & 12.09 & 0.28 & 374.95 & 18.27 & 0.63 & 1329.44\\
    \midrule
    \multirow{3}{*}{metrics} & $LR$ & 0 & 0 & $9.9481\times 10 ^{-7}$ & 0 & 0 & 0 & 0\\ %\cmidrule{2-9}
    & $Kurt$ & 11618.84 & 1926.231 & 81.92 & 12632.84 & 2388.49 & 101.9082 & 34770\\ %\cmidrule{2-9}
    & $CORR$ & 0.99984 & 0.99861 & 0.97295 & 0.99982 & 0.99929 & 0.98486 & 0.99999\\ 
    \bottomrule
    \end{tabular}
 \end{table*}

\subsection{Additional Studies}
\jq{
\subsubsection{Threshold}
\label{sssec:threshold}
This section calculates the $TPR$ (True Positive Rate) and $FPR$ (False Positive Rate) of the similarity distance to determine the distance threshold. 
%allowing for a weighty reference for our method.
$TPR$ is the probability of perfectly detecting the duplicate and the source model, and $FPR$ is the probability of mistaking a non-duplicate as a copy of the source version.
% and $FPR$ is the probability of mistaking the non-duplicate and the source version.
To this end, we calculate the $TPR$ and $FPR$ of the HM25 data set. Firstly, we obtain the positive sample set $\mathcal{P}$ and the negative sample set $\mathcal{N}$, where $\mathcal{P}=\{(X,Y)|C(X,Y)=1,X \in HM25, Y \in HM25\}$, $\mathcal{N}=\{(X,Y) | C(X,Y)=0,X \in HM25, Y \in HM25\}$.  $C(X,Y)=1$ denotes that $X$ is the copy of $Y$; otherwise, they are not a copy relationship. Note that the subdivision attack is not considered here. 
Subsequently, we calculate the similarity distance sets $\mathcal{PD}$ and $\mathcal{ND}$ of the model pairs in $\mathcal{P}$ and $\mathcal{N}$, respectively, and compute the $TPR$ and $FPR$ of the $\mathcal{PD}$ and $\mathcal{ND}$ according to multiple preset thresholds. 
Given threshold $t$, $TPR$ and $FPR$ of LR distance and CORR distance are calculated as:
\begin{equation}
\begin{aligned}
\label{eq:tpr}
    TPR_{LR} &= \frac{|\{x|x\le t,x\in \mathcal{PD}\}|}{|\mathcal{PD}|},\\
    FPR_{LR} &= \frac{|\{x|x\le t,x\in \mathcal{ND}\}|}{|\mathcal{ND}|},\\
    TPR_{CORR} &= \frac{|\{x|x\ge t, x\in \mathcal{PD}\}|}{|\mathcal{PD}|},\\
    FPR_{CORR} &= \frac{|\{x|x\ge t,x\in \mathcal{ND}\}|}{|\mathcal{ND}|},\\
\end{aligned}
\end{equation}
where $|\cdot|$ denotes the number of elements in the set. 
Ultimately we decide the optimal threshold in line with maximizing $TPR-FPR$. Figure  \ref{fig:threshold} illustrates the trend of $TPR_{LR}$, $FPR_{LR}$, $TPR_{CORR}$, and $FPR_{CORR}$ with varying the threshold. 
When the threshold is $0.9951$, $(TPR_{CORR},\\FPR_{CORR})=(0.9259, 0.0369)$, and when the threshold is $0.0006$, $(TPR_{LR},FPR_{LR}) = (0.8365,0.0413)$, maximizing the difference between $TPR$ and $FPR$. In this way, we empirically find the optimal threshold of CORR distance $t_{CORR}=0.9951$, and the optimal threshold of LR distance $t_{LR}=0.0006$. 
\begin{figure}
    \centering
    \begin{minipage}[b]{0.47\linewidth}
    \subfigure[threshold of CORR]{\label{fig:threshold_corr}\includegraphics[width=1\linewidth]{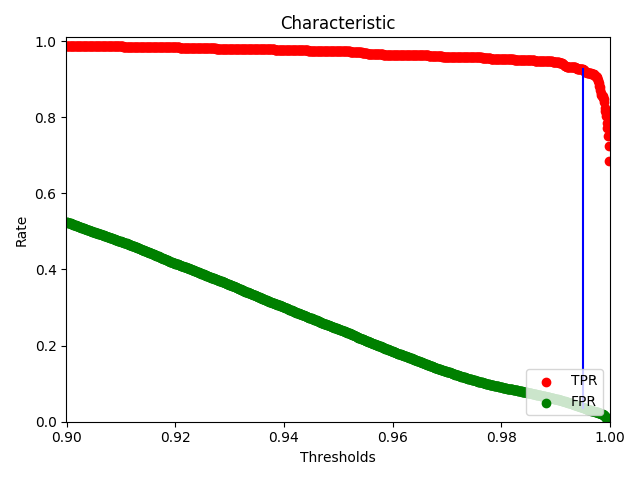}}
    \end{minipage}
    \begin{minipage}[b]{0.47\linewidth}
    \subfigure[threshold of LR]{\label{fig:threshold_LR}\includegraphics[width=1\linewidth]{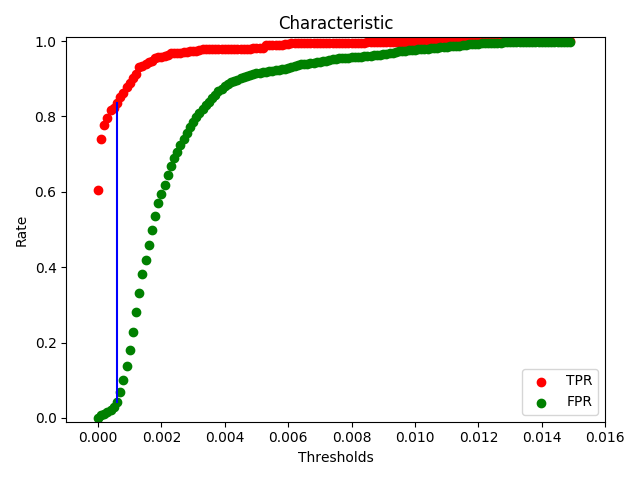}}
    \end{minipage}
    \caption{The curve of $TPR$ and $FPR$ of LR and CORR varying with the threshold. The blue line indicates the maximum $TPR-FPR$ under this threshold.}
    \label{fig:threshold}
\end{figure}
}

\subsubsection{Downsampling}
In this section, we prove that downsampling improves the efficiency of point set registration. To this end, we choose the Bunny model and its $5\%$ noisy version as the experimental objects, both of which have 34,835 points and 69,666 faces, and conduct the following experiments. 
Firstly, we calculate the measures between the two models as the baseline. We then perform HEM downsampling \cite{preiner2014continuous} and random downsampling on the two models to obtain point clouds with different numbers of points. The downsampling parameters are adjusted to ensure the same number of points in both sampling schemes. 
Finally, we calculate the similarity distances and registration runtime of the point clouds above.

Table \ref{table:Downsampling} shows the effectiveness of the downsampling strategy. It is observed that the higher the sampling rate is, the greater the advantage of HEM over random downsampling appears. Downsampling reduces the running time of the CPD algorithm (i.e., point set registration) immensely because of the plunge of point number. Meanwhile, it has little impact on LR and CORR. % All three distances manifest the role of HEM in narrowing the gap. 
The maximum modification of only $10^{-7}$ indicates the best robustness of the LR measure against downsampling. The CORR measure varies more than the LR distance with a tiny difference of $0.03$. Kurt is sensitive to downsampling (i.e., number of points), but it is still doable in judging models with the same number of points.

Table \ref{table:HEM} compares the runtime of HEM and random downsampling. It is feasible to control the approximate sampling rates for the Horse model with 112,642 points and the Bunny model with 34,835 points. In general, the HEM algorithm demands more extra time than the random downsampling. However, HEM is still fast and occupies a tiny portion, considering point set registration. 

\begin{table}[thbp]
    \caption{Runtime of HEM and random downsampling. }
    \label{table:HEM}
    \centering
    \begin{tabular}{c c c c c}
    \toprule
     & \multicolumn{2}{c}{Horse} & \multicolumn{2}{c}{Bunny} \\
    \midrule
        layers & 5 & / & 5 & /\\
        amendatory factor & 12 & / & 6.7 & /\\
        sampling rate ($\%$) & / & 0.471 & / & 1.521 \\
        size  & $532\times3$ & $530\times3$ & $546\times3$ & $530\times3$\\
        time(s) & 10.48 & 0.03 & 1.6359 & 0.02\\
    \bottomrule
    \end{tabular}
\end{table}

\subsubsection{Segmentation}
In this section, we evaluate the impact of the segmentation strategy on the LR measure. The Merlion model with 17,705 points and 35,414 faces and its homologous model after similar transformation are used for the experiments. The registered point clouds are first divided into multiple segments, and the LR distance is then computed for each segments pair. Finally, we merge the LR results of each segment and compare the effects of segment numbers on our method.

Table \ref{table:Segmentation} reveals that more segments would lead to less time and memory but hardly impact LR distance. $>x$ means that the value fluctuates around $x$ within a stable and observable duration. The Merlion models in this experiment achieve good alignment through CPD, and thus the time required for the IALM operation is relatively short without segmentation strategy. The memory is related to the size of the input matrix of the IALM algorithm, and the running time of the IALM is related to the complexity of the matrix.

\begin{table}[thbp]\tablefont
    \centering
    \caption{Evaluation of segmentation strategy on our method. 
    The points item shows the number of points in the first $(n-1)$ block / the number of points in the last block. } 
    \label{table:Segmentation}
    \begin{tabular}{c c c c c}
    \toprule
    \multirow{2}{*}{} & \multicolumn{4}{c}{Merlion} \\  \cmidrule{2-5}
    threshold & base & 3000 & 4000 & 5000\\
    \midrule
    blocks & 1 & 6 & 5 & 4\\
    points & 17705 & 2950/2955 & 3541/3541 & 4426/4427\\
    IALM time(s)& 13.3664  & 4.14514 & 4.03534 & 4.5393 \\
    memory & $>20$G & $>1$G & $>1$G & $>1$G \\
    LR & 0 & 0 & 0 & 0\\
    \bottomrule
    \end{tabular}
\end{table}

\section{Conclusion}
In this paper, we presented a robust method for copyrighting 3D point cloud data. It first registers two point clouds and then computes different distance measures between them. The distance measures reveal the similarity degree of the two point clouds, enabling the judgement of the similar models or two different models. Extensive experiments show that our method generally achieves better outcomes than the state-of-the-art watermarking techniques and comparable performance to current 3D shape retrieval methods. We believe our work will inspire more insights in terms of copyrighting 3D shapes.

One main limitation of our work is the efficiency due to the complex computation in point set registration. In other words, our method is more suitable for offline processing. In the future, we would like to design efficient techniques to enable fast processing.
% -------------------------

% Numbered list
% Use the style of numbering in square brackets.
% If nothing is used, default style will be taken.
%\begin{enumerate}[a)]
%\item 
%\item 
%\item 
%\end{enumerate}  

% Unnumbered list
%\begin{itemize}
%\item 
%\item 
%\item 
%\end{itemize}  

% Description list
%\begin{description}
%\item[]
%\item[] 
%\item[] 
%\end{description}  

% Figure
% \begin{figure}[<options>]
% 	\centering
% 		\includegraphics[<options>]{}
% 	  \caption{}\label{fig1}
% \end{figure}

% \begin{table}[<options>]
% \caption{}\label{tbl1}
% \begin{tabular*}{\tblwidth}{@{}LL@{}}
% \toprule
%   &  \\ % Table header row
% \midrule
%  & \\
%  & \\
%  & \\
%  & \\
% \bottomrule
% \end{tabular*}
% \end{table}

% Uncomment and use as the case may be
%\begin{theorem} 
%\end{theorem}

% Uncomment and use as the case may be
%\begin{lemma} 
%\end{lemma}

%% The Appendices part is started with the command \appendix;
%% appendix sections are then done as normal sections
%% \appendix

% \section{}\label{}

% To print the credit authorship contribution details
% \printcredits

%% Loading bibliography style file
%\bibliographystyle{model1-num-names}
\bibliographystyle{cas-model2-names}

% Loading bibliography database
\bibliography{cas-refs.bib}

% Biography
\bio{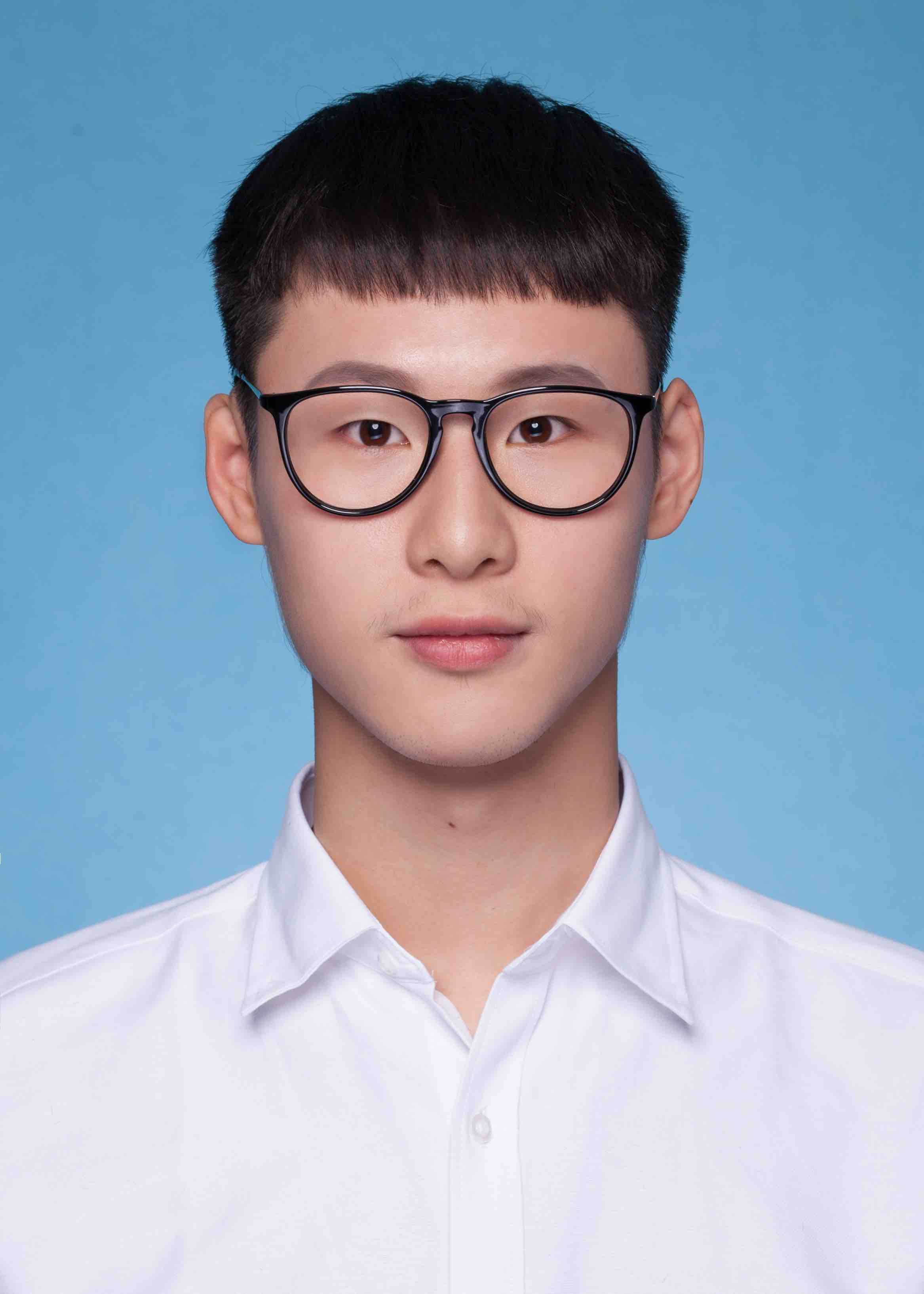}
Jiaqi Yang received a master's degree in computer science and technology from the School of Computer Science, Zhejiang University. His current research interests include computer graphics and other fields.
\endbio

\vspace{25mm}

% \newpage
\bio{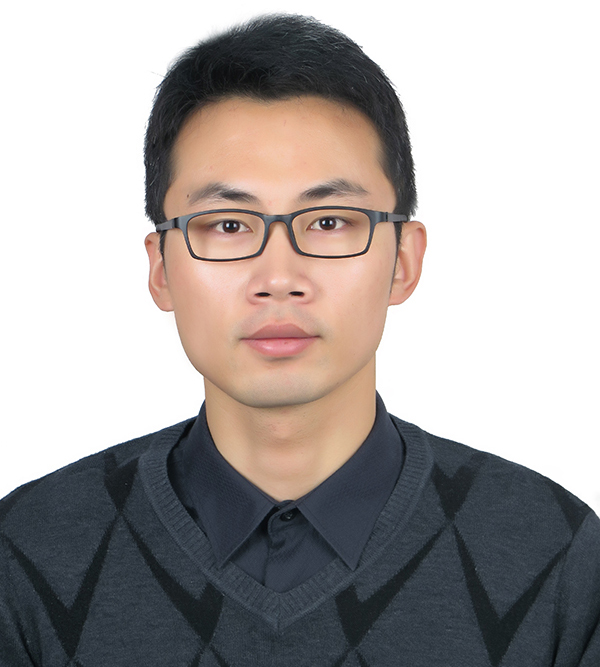}
Xuequan Lu is an Assistant Professor at the School of Information Technology, Deakin University, Australia. He spent more than two years as a Research Fellow in Singapore. Before that, he earned his Ph.D. at Zhejiang University (China) in June 2016. His research interests mainly fall into visual computing, for example, geometry modeling, processing and analysis, animation/simulation, 2D data processing, and analysis. More information can be found at http://www.xuequanlu.com.
\endbio
\vspace{25mm}
 \newpage
\bio{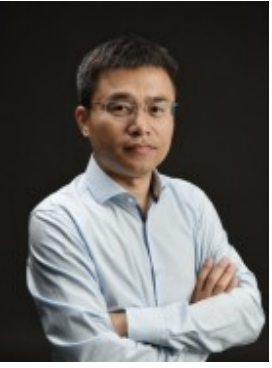}
Wenzhi Chen (Member, IEEE) received a Ph.D. degree from the College of Computer Science and Engineering, Zhejiang University. He is currently a Professor with the College of Computer Science and Technology, Zhejiang University, and the Director of the Information Technology Center, Zhejiang University. He used to be the Vice Dean of the College of Computer Science and Technology. His current research interests include embedded systems and their application, computer architecture, computer system software, and information security. He is a member of ACM and the ACM Education Council.
\endbio
\end{document}